# Semantic Matchmaking as Non-Monotonic Reasoning: A Description Logic Approach

**Tommaso Di Noia**                                    T.DINOIA@POLIBA.IT
**Eugenio Di Sciascio**                               DISCIASCIO@POLIBA.IT
*SisInfLab - Politecnico di Bari, Bari, Italy*

**Francesco M. Donini**                                DONINI@UNITUS.IT
*Università della Tuscia, Viterbo, Italy*

## Abstract

Matchmaking arises when supply and demand meet in an electronic marketplace, or when agents search for a web service to perform some task, or even when recruiting agencies match curricula and job profiles. In such open environments, the objective of a matchmaking process is to discover best available offers to a given request.

We address the problem of matchmaking from a knowledge representation perspective, with a formalization based on Description Logics. We devise Concept Abduction and Concept Contraction as non-monotonic inferences in Description Logics suitable for modeling matchmaking in a logical framework, and prove some related complexity results. We also present reasonable algorithms for semantic matchmaking based on the devised inferences, and prove that they obey to some commonsense properties.

Finally, we report on the implementation of the proposed matchmaking framework, which has been used both as a mediator in e-marketplaces and for semantic web services discovery.

## 1. Introduction

The promise of the Semantic Web initiative is to revolutionize the way information is coded, stored, and searched on the Internet (Berners-Lee, Hendler, & Lassila, 2001). The basic idea is to structure information with the aid of markup languages, based on the XML language, such as RDF and RDFS[1], and OWL[2]. These languages have been conceived for the representation of machine-understandable, and unambiguous, description of web content through the creation of domain ontologies, and aim at increasing openness and interoperability in the web environment.

Widespread availability of resources and services enables—among other advantages— the interaction with a number of potential counterparts. The bottleneck is that it is difficult finding matches, possibly the best ones, between parties.

The need for a matchmaking process arises when supply and demand have to meet in a marketplace, or when web services able to perform some task have to be discovered, but also when recruiting agencies match curricula and job profiles or a dating agency has to propose partners to a customer of the agency. Requests and offers may hence be generic demands and supplies, web services, information, tangible or intangible goods, and a matchmaking process should find for any request an appropriate response. In this paper we concentrate on automated matchmaking, basically oriented to electronic marketplaces and service discovery, although principles and algorithms are definitely general enough to cover also other scenarios. We assume, as it is reasonable, that both requests and offers are endowed of some kind of description. Based on these descriptions the target of the matching process is finding, for a given request, best matches available in the offers set, and also, given an offer, determine best matching requests in a peer-to-peer fashion. We may hence think of an electronic mediator as the actor who actively tries to carry out the matchmaking process. Obviously descriptions might be provided using unstructured text, and in this case such an automated mediator should revert to adopting either basic string matching techniques or more sophisticated Information Retrieval techniques.

---

1. http://www.w3.org/RDF/
2. http://www.w3.org/TR/owl-features/

The Semantic Web paradigm calls for descriptions that should be provided in a structured form based on ontologies, and we will assume in what follows that requests and offers are given with reference to a common ontology. It should be noticed that even when requests and offers are described in heterogeneous languages, or using different ontologies modelling the same domain, schema/data integration techniques may be employed to make them comparable, as proposed *e.g.*, by Madhavan, Bernstein, and Rahm (2001), and Shvaiko and Euzenat (2005); but once they are reformulated in a comparable way, one is still left with the basic matchmaking problems: given a request, are there compatible offers? If there are several compatible offers, which, and why, are the most promising ones?

Matchmaking has been widely studied and several proposals have been made in the past; we report on them in Section 2. Recently, there has been a growing effort aimed at the formalization with Description Logics (DLs) (Baader, Calvanese, Mc Guinness, Nardi, & Patel-Schneider, 2003) of the matchmaking process (*e.g.*, Di Sciascio, Donini, Mongiello, & Piscitelli, 2001; Trastour, Bartolini, & Priest, 2002; Sycara, Widoff, Klusch, & Lu, 2002; Di Noia, Di Sciascio, Donini, & Mongiello, 2003b; Li & Horrocks, 2003; Di Noia, Di Sciascio, Donini, & Mongiello, 2003c, 2003a, among others). DLs, in fact, allow to model structured descriptions of requests and offers as concepts, usually sharing a common ontology. Furthermore DLs allow for an open-world assumption. Incomplete information is admitted, and absence of information can be distinguished from negative information. We provide a little insight on DLs in Section 3.

Usually, DL-based approaches exploit standard reasoning services of a DL system—subsumption and (un)satisfiability—to match potential partners in an electronic transaction. In brief, if a supply is described by a concept *Sup* and a demand by a concept *Dem*, unsatisfiability of the conjunction of *Sup* and *Dem* (noted as $Sup \sqcap Dem$) identifies the incompatible proposals, satisfiability identifies potential partners—that still have to agree on underspecified constraints—and subsumption between *Sup* and *Dem* (noted as $Sup \sqsubseteq Dem$) means that requirements on *Dem* are completely fulfilled by *Sup*.

Classification into compatible and incompatible matches can be useless in the presence of several compatible supplies; some way to rank most promising ones has to be identified; also some explanation on motivation of such a rank could be appreciated. On the other hand, when there is lack of compatible matches one may accept to turn to incompatible matches that could still be interesting, by revising some of the original requirements presented in the request, as far as one could easily identify them.

In other words some method is needed to provide a logic-based score for both compatible and incompatible matches and eventually provide a partial/full ordering, allowing a user or an automated agent to choose most promising counteroffers. Furthermore it should be possible, given a score, to provide logical explanations of the resulting score, thus allowing to understand the rank result and ease further interaction to refine/revise the request. Although this process is quite simple for a human being it is not so in a logic-based fully automated framework. We believe there is a need to define non-monotonic reasoning services in a DLs setting, to deal with approximation and ranking, and in this paper we propose the use of Concept Abduction (Di Noia et al., 2003a) and Concept Contraction (Colucci, Di Noia, Di Sciascio, Donini, & Mongiello, 2003), as services amenable to answer the above highlighted issues in a satisfactory way. Contributions of this paper include:

- a logical framework to express requests and offers in terms of concept descriptions, and properties that should hold in a matchmaking facilitator;

- Concept Abduction as a logical basis for ranking compatible counteroffers to a given offer and provide logical explanations of the ranking result;

- Concept Contraction as a logical basis for ranking incompatible matches, aimed at discovering most promising "near misses", and provide logical explanations of the ranking result;

- algorithms implementing the formalized inferences for matchmaking purposes and complexity results for a class of matchmaking problems;

- description of our system implementing semantic matchmaking services, and experimental evaluation.

The remaining of the paper is structured as follows: next Section reports on background work on the subject. Then (Section 3) we briefly revise Description Logics basics. To make the paper self-contained we recall (Section 4) our logic-based framework for matchmaking, pointing out properties that matchmaking algorithms and systems should guarantee. In Sections 5 and 6 we present Concept Abduction and Concept Contraction, the two inference services we devised to compute semantic matchmaking, and present suitable definitions of the problem along with some complexity results. Then in Section 7 we describe our matchmaker, and present (Section 7.1) an evaluation of results computed by the system compared with human users behavior, and with a standard full text retrieval approach. Conclusions close the paper.

## 2. Related Work on Matchmaking

Matchmaking has been investigated in recent years under a number of perspectives and for different purposes, with a renovated interest as the information overload kept growing with the Web widespreading use. We try here to summarize some of the relevant related work. Vague query answering, proposed by Motro (1988), was an initial effort to overcome limitations of relational databases, using weights attributed to several search variables. More recent approaches along these lines aim at extending SQL with "preference" clauses, in order to softly matchmake data in structured databases (Kießling, 2002). Finin, Fritzson, McKay, and McEntire (1994) proposed KQML as an agent communication language oriented to matchmaking purposes. Kuokka and Harada (1996) investigated matchmaking as a process that allowed potential producers/consumers to provide descriptions of their products/needs, either directly or through agents mediation, to be later unified by an engine identifying promising matches. Two engines were developed, the SHADE system, which again used KQML, and as description language KIF, with matchmaking anyway not relying on any logical reasoning, and COINS, which adopted classical unstructured-text information retrieval techniques, namely the SMART IR system. Similar methods were later re-considered in the GRAPPA system (Veit, Muller, Schneider, & Fiehn, 2001). Classified-ads matchmaking, at a syntactic level, was proposed by Raman, Livny, and Solomon (1998) to matchmake semi-structured descriptions advertising computational resources in a fashion anticipating Grid resources brokering. Matchmaking was used in SIMS (Arens, Knoblock, & Shen, 1996) to dynamically integrate queries; the approach used KQML, and LOOM as description language. LOOM is also used in the subsumption matching addressed by Gil and Ramachandran (2001). InfoSleuth (Jacobs & Shea, 1995), a system for discovery and integration of information, included an agent matchmaker, which adopted KIF and the deductive database language LDL++. Constraint-based approaches to matchmaking have been proposed and implemented in several systems, $e.g.$, PersonaLogic[3], Kasbah[4] and systems by Maes, Guttman, and Moukas (1999), Karacapilidis and Moraitis (2001), Wang, Liao, and Liao (2002), Ströbel and Stolze (2002).

Matchmaking as satisfiability of concept conjunction in DLs was first proposed in the same venue by Gonzales-Castillo, Trastour, and Bartolini (2001) and by Di Sciascio et al. (2001), and precisely defined by Trastour et al. (2002). Sycara, Paolucci, Van Velsen, and Giampapa (2003) introduced a specific language for agent advertisement in the framework of the Retsina Multiagent infrastructure. A matchmaking engine was developed (Sycara et al., 2002; Paolucci, Kawamura, Payne, & Sycara, 2002), which carries out the process on five possible levels. Such levels exploit both classical text-retrieval techniques and semantic match using Θ-subsumption. Nevertheless, standard features of a semantic-based system, as satisfiability check are unavailable. It is noteworthy that in this approach, the notion of *plug-in* match is introduced, to overcome in some way the limitations of a matching approach based on exact matches. The approach of Paolucci et al. (2002) was later extended by Li and Horrocks (2003), where two new levels for matching classification were introduced. A similar classification was proposed—in the same venue—by Di Noia et al. (2003c), along with properties that a matchmaker should have in a DL-based framework, and algorithms to classify and semantically rank matches within classes. Benatallah, Hacid, Rey, and Toumani (2003) proposed the Difference Operator in DLs for semantic matchmaking. The approach uses Concept Difference, followed by a covering operation optimized using hypergraph techniques, in the framework of web services discovery. We briefly comment on the relationship between Concept Difference and Concept Abduction at the end of Section 5. An initial DL-based approach, adopting penalty functions ranking, has been proposed by Calì, Calvanese, Colucci, Di Noia, and Donini (2004), in the framework of dating systems. An extended matchmaking approach, with negotiable and strict constraints in a DL framework has been proposed by Colucci, Di Noia, Di Sciascio, Donini, and Mongiello (2005), using both Concept Contraction and Concept Abduction. Matchmaking in DLs with locally-closed world

---

3. http://www.PersonaLogic.com

4. http://www.kasbah.com

assumption applying autoepistemic DLs has been proposed by Grimm, Motik, and Preist (2006).

The need to work in someway with approximation and ranking in DL-based approaches to matchmaking has also recently led to adopting fuzzy-DLs, as in Smart (Agarwal & Lamparter, 2005) or hybrid approaches, as in the OWLS-MX matchmaker (Klusch, Fries, Khalid, & Sycara, 2005). Such approaches, anyway, relaxing the logical constraints, do not allow any explanation or automated revision service.

Finally, it should be pointed out that *matching* in DLs, widely treated by Baader, Küsters, Borgida, and Mc Guinness (1999) has no relation to matchmaking. In fact, in that work expressions denoting concepts are considered, with variables in expressions. Then a match is a substitution of variables with expressions that makes a concept expression equivalent to another. Also the more general setting of concept rewriting in DLs has no direct relation with matchmaking—see the discussion in Remark 1.

## 3. Description Logics Basics

In this Section we summarize the basic notions and definitions about Description Logics (DLs), and about Classic, the knowledge representation system our application is inspired by. We provide hereafter a brief guided-tour of DLs main characteristics, while the interested reader can refer to the comprehensive handbook by Baader et al. (2003).

### 3.1 Description Logics

Description Logics—a.k.a. Terminological Logics—are a family of logic formalisms for Knowledge Representation. All DLs are endowed of a syntax, and a semantics, which is usually model-theoretic. The basic syntax elements of DLs are:

- *concept* names, *e.g.*, `Computer`, `CPU`, `Device`, `Software`,

- *role* names, like `hasSoftware`, `hasDevice`

- *individuals*, that are used for special named elements belonging to concepts.

Intuitively, concepts stand for sets of objects, and roles link objects in different concepts, as the role `hasSoftware` that links computers to software. We are not using individuals in our formalization, hence from now on we skip the parts regarding individuals.

Formally, a semantic *interpretation* is a pair $\mathcal{I} = (\Delta, \cdot^{\mathcal{I}})$, which consists of the *domain* $\Delta$ and the *interpretation function* $\cdot^{\mathcal{I}}$, which maps every concept to a subset of $\Delta$, and every role to a subset of $\Delta \times \Delta$.

Basic elements can be combined using *constructors* to form concept and role *expressions*, and each DL has its distinguished set of constructors. Every DL allows one to form a *conjunction* of concepts, usually denoted as $\sqcap$; some DL include also disjunction $\sqcup$ and complement $\neg$ to close concept expressions under boolean operations.

Roles can be combined with concepts using

- *existential role quantification*:

  *e.g.*, `Computer` $\sqcap \exists$`hasSoftware.WordProcessor`

  which describes the set of computers whose software include a word processor, and

- *universal role quantification*

  *e.g.*, `Server` $\sqcap \forall$`hasCPU.Intel`

  which describes servers with only Intel processors on board.

  Other constructs may involve counting, as

- *number restrictions*:

  *e.g.*, `Computer` $\sqcap (\leq 1$ `hasCPU`$)$

  expresses computers with at most one CPU, and

  *e.g.*, `Computer` $\sqcap (\geq 4$ `hasCPU`$)$

  describes computers equipped with at least four CPUs.

Many other constructs can be defined, increasing the expressive power of the DL, up to n-ary relations (Calvanese, De Giacomo, & Lenzerini, 1998).

In what follows, we call *atomic concepts* the union of concept names, negated concept names, and unqualified number restrictions. We define *length* of a concept $C$ as the number of atomic concepts appearing in $C$. We denote the length of $C$ as $|C|$. Observe that we consider $\top$ and $\bot$ to have zero length. We define the Quantification Nesting (QN) of a concept as the following positive integer: the QN of an atomic concept is 0, the QN of a universal role quantification $\forall R.F$ is 1 plus the QN of $F$, and the QN of a conjunction $C_1 \sqcap C_2$ is the maximum between the QNs of conjoined concepts $C_1$ and $C_2$.

Expressions are given a semantics by defining the interpretation function over each construct. For example, concept conjunction is interpreted as set intersection: $(C \sqcap D)^{\mathcal{I}} = C^{\mathcal{I}} \cap D^{\mathcal{I}}$, and also the other boolean connectives $\sqcup$ and $\neg$, when present, are given the usual set-theoretic interpretation of union and complement. The interpretation of constructs involving quantification on roles needs to make domain elements explicit: for example, $(\forall R.C)^{\mathcal{I}} = \{d_1 \in \Delta \mid \forall d_2 \in \Delta : (d_1, d_2) \in R^{\mathcal{I}} \rightarrow d_2 \in C^{\mathcal{I}}\}$

### 3.2 TBoxes

Concept expressions can be used in *axioms*—that can be either *inclusions* (symbol: $\sqsubseteq$), or *definitions* (symbol: $\equiv$)—which impose restrictions on possible interpretations according to the knowledge elicited for a given domain. For example, we could impose that monitors can be divided into CRT and LCD using the two inclusions: `Monitor` $\sqsubseteq$ `LCDMonitor` $\sqcup$ `CRTMonitor` and `CRTMonitor` $\sqsubseteq \neg$`LCDMonitor`. Or, that computers for a domestic use have only one operating system as `HomePC` $\sqsubseteq (\leq 1$ `hasOS`$)$. Definitions are useful to give a meaningful name to particular combinations, as in `Server` $\equiv$ `Computer` $\sqcap (\geq 2$ `hasCPU`$)$.

Historically, sets of such axioms are called a TBox (Terminological Box). There are several possible types of TBoxes. General TBoxes are made by General Concept Inclusions (GCI) of the form $C \sqsubseteq D$, where both $C$ and *Dem* can be any concept of the DL. For general TBoxes, the distinction between inclusions and definitions disappears, since any definition $C \equiv D$ can be expressed by two GCIs $C \sqsubseteq D, D \sqsubseteq C$. On the contrary, in *simple* TBoxes—also called *schemas* by Calvanese (1996), and by Buchheit, Donini, Nutt, and Schaerf (1998)—only a concept name can appear on the left-hand side (l.h.s.) of an axiom, and a concept name can appear on the l.h.s. of at most one axiom. Schemas can be cyclic or acyclic, where cyclicity refers to the *dependency graph* $G_\mathcal{T}$ between concept names, defined as follows: every concept name is a node in $G_\mathcal{T}$, and there is an arc from concept name $A$ to concept name $B$ if $A$ appears on the l.h.s. of an axiom, and $B$ appears (at any level) in the concept on the right-hand side. $\mathcal{T}$ is acyclic if $G_\mathcal{T}$ is, and it is cyclic otherwise. We call an acyclic schema a *simple* TBox (Baader et al., 2003, Ch.2). The *depth* of a simple TBox $\mathcal{T}$ is the length of the longest path in $G_\mathcal{T}$. Only for simple TBoxes, *unfolding* has been defined as the following process (see Appendix A for a definition): for every definition $A \equiv C$, replace $A$ with $C$ in every concept; for every inclusion $A \sqsubseteq C$, replace $A$ with $A \sqcap C$ in every concept. Clearly, such a process trasforms every concept into an equivalent one, where the TBox can be forgotten. However, for some TBoxes, unfolding can yield concepts of exponential size w.r.t. the initial concepts. When such an exponential blow-up does not happen, we call the TBox "bushy but not deep" (Nebel, 1990).

The semantics of axioms is based on set containment and equality: an interpretation $\mathcal{I}$ satisfies an inclusion $C \sqsubseteq D$ if $C^\mathcal{I} \subseteq D^\mathcal{I}$, and it satisfies a definition $C \equiv D$ when $C^\mathcal{I} = D^\mathcal{I}$. A *model* of a TBox $\mathcal{T}$ is an interpretation satisfying all axioms of $\mathcal{T}$.

Observe that we make a distinction between equivalence $\equiv$ (used in axioms) and equality $=$ symbols. We use equality to instantiate generic concept symbols with the concepts they stand for, *e.g.*, when we write "... where $C = A \sqcap \forall R.B$..." we mean that the concept symbol $C$ stands for the concept expression $A \sqcap \forall R.B$ in the text.

### 3.3 Reasoning Services

DL-based systems usually provide two basic reasoning services:

1. *Concept Satisfiability*: given a TBox $\mathcal{T}$ and a concept $C$, does there exist at least one model of $\mathcal{T}$ assigning a non-empty extension to $C$? We abbreviate satisfiability of a concept $C$ w.r.t. a TBox $\mathcal{T}$ as $C \not\sqsubseteq_\mathcal{T} \bot$.

2. *Subsumption*: given a TBox $\mathcal{T}$ and two concepts $C$ and $D$, is $C^\mathcal{I}$ always contained in $D^\mathcal{I}$ for every model $\mathcal{I}$ of $\mathcal{T}$? We abbreviate subsumption between $C$ and $D$ w.r.t. $\mathcal{T}$ as $C \sqsubseteq_\mathcal{T} D$.

Since $C$ is satisfiable iff $C$ is not subsumed by $\bot$, complexity lower bounds for satisfiability carry over (for the complement class) to subsumption, and upper bounds for subsumption carry over to satisfiability. On the other hand, since $C$ is subsumed by $D$ iff $C \sqcap \neg D$ is unsatisfiable, subsumption is reducible to satisfiability in DLs admitting general concept negation, but not in those DLs in which $\neg D$ is outside the language—as in the DLs of the next Section.

### 3.4 The System Classic

The system Classic (Borgida, Brachman, McGuinness, & A. Resnick, 1989; Borgida & Patel-Schneider, 1994) has been originally developed as a general Knowledge Representation system, and has been successfully applied to configuration (Wright, Weixelbaum, Vesonder, Brown, Palmer, Berman, & Moore, 1993) and program repositories management (Devambu, Brachman, Selfridge, & Ballard, 1991).

Its language has been designed to be as expressive as possible while still admitting polynomial-time inferences for "bushy but not deep" TBoxes. So it provides intersection of concepts but no union, universal but not existential quantification over roles, and number restrictions over roles but no intersection of roles, since each of these combinations is known to make reasoning NP-hard (Donini, Lenzerini, Nardi, & Nutt, 1991; Donini, 2003).

| name | concrete syntax | syntax | semantics |
|---|---|---|---|
| top | TOP | $\top$ | $\Delta^{\mathcal{I}}$ |
| bottom | - | $\bot$ | $\emptyset$ |
| intersection | (and C D) | $C \sqcap D$ | $C^{\mathcal{I}} \cap D^{\mathcal{I}}$ |
| universal quantification | (all R C) | $\forall R.C$ | $\{d_1 \mid \forall d_2 : (d_1, d_2) \in R^{\mathcal{I}} \rightarrow d_2 \in C^{\mathcal{I}}\}$ |
| number restrictions | (at-least n R) | $(\geq n\ R)$ | $\{d_1 \mid \sharp\{d_2 \mid (d_1, d_2) \in R^{\mathcal{I}}\} \geq n\}$ |
| | (at-most n R) | $(\leq n\ R)$ | $\{d_1 \mid \sharp\{d_2 \mid (d_1, d_2) \in R^{\mathcal{I}}\} \leq n\}$ |

Table 1: Syntax and semantics of some constructs of CLASSIC

| name | system notation | syntax | semantics |
|---|---|---|---|
| definition | (createConcept A C false) | $A \equiv C$ | $A^{\mathcal{I}} = C^{\mathcal{I}}$ |
| inclusion | (createConcept A C true) | $A \sqsubseteq C$ | $A^{\mathcal{I}} \subseteq C^{\mathcal{I}}$ |
| disjoint group | (createConcept $A_1$ C *symbol*) ... (createConcept $A_k$ C *symbol*) | $disj(A_1, \ldots, A_k)$ | for $i = 1, \ldots, k$ $A_i^{\mathcal{I}} \subseteq C^{\mathcal{I}}$ and for $j = i + 1, \ldots, k$ $A_i^{\mathcal{I}} \cap A_j^{\mathcal{I}} = \emptyset$ |

Table 2: Syntax and semantics of the TBox CLASSIC assertions (*symbol* is a name denoting the group of disjoint concepts)

For simplicity, we only consider a subset of the constructs, namely, conjunction, number restrictions, and universal role quantifications, summarized in Table 1. We abbreviate the conjunction $(\geq n\ R) \sqcap (\leq n\ R)$ as $(= n\ R)$. We omit constructs ONE-OF$(\cdot)$, FILLS$(\cdot, \cdot)$ that refer to individuals, and construct SAME-AS$(\cdot, \cdot)$ equating fillers in functional roles. The subset of CLASSIC we refer to is known as $\mathcal{ALN}$ (Attributive Language with unqualified Number restrictions) (Donini, Lenzerini, Nardi, & Nutt, 1997b). When number restrictions are not present, the resulting DL is known as $\mathcal{AL}$ (Schmidt-Schauß & Smolka, 1991). $\mathcal{ALN}$ provides a minimal set of constructs that allow one to represent a concept taxonomy, disjoint groups, role restrictions ($\mathcal{AL}$), and number restrictions ($\mathcal{N}$) to represent restriction son the number of fillers of a role.

Regarding axioms in a TBox, CLASSIC allows one to state a simple TBox of assertions of the form summarized in Table 2, where $A, A_1, \ldots, A_k$ are all concept names. Axioms in the TBox are subject to the constraints that every concept name can appear at most once as the l.h.s. in a TBox, and every concept name cannot appear both on the l.h.s. of a definition and in a disjointness assertion.

Every CLASSIC concept can be given a *normal form.* Here we consider the normal form only for the constructs of $\mathcal{ALN}$ that we used in the ontologies and applications. Intuitively, the normal form pre-computes all implications of a concept, including—possibly—its unsatisfiability. The normal form can be reached, up to commutativity of the operator $\sqcap$, using well-known normalization rules, that we report in Appendix A to make the paper self-contained. The normal form of an unsatisfiable concept is simply $\bot$. Every satisfiable concept $C$ can be divided into three components: $C_{names} \sqcap C_\sharp \sqcap C_{all}$. The component $C_{names}$ is the conjunction of all concept names $A_1, \ldots, A_h$. The component $C_\sharp$ is the conjunction of all number restrictions, no more than two for every role (the maximum at-least and the minimum at-most for each role), including for every conjunct of $C$ of the form $\forall R.\bot$, the number restriction $(\leq 0 R)$ in $C_\sharp$. The component $C_{all}$ conjoins all concepts of the form $\forall R.D$, one for each role $R$, where $D$ is again in normal form. We call such form Conjunctive Normal Form—CNF, in analogy with Propositional Logic—and we observe that CNF is unique (also said *canonical*), up to commutativity of conjunction.

Moreover, the TBox in Classic can be embedded in the concepts, by expanding definitions, and adding the right-hand side concepts of inclusions, and adding the negation of disjoint concept names—see Appendix A for more details. For instance, suppose that a TBox contains:

1. the definition `Server` $\equiv$ `Computer` $\sqcap$ $(\geq 2 \text{ hasCPU})$,

2. the inclusion `Computer` $\sqsubseteq$ $(\geq 1 \text{ hasStorageDevice})$,

3. and the disjointness assertion $disj(\text{AMD}, \text{Intel})$.

Then, the concept `Server`$\sqcap\forall$`hasCPU.Intel` can be rewritten into `Computer`$\sqcap(\geq 2 \text{ hasCPU})\sqcap$ $(\geq 1 \text{ hasStorageDevice})\sqcap\forall$`hasCPU.`$(\text{Intel}\sqcap\neg\text{AMD})$, which is equivalent to the former w.r.t. models of the TBox. Observe that the concept name `Computer` is kept in the rewriting, since the inclusion gives only a necessary condition $(\geq 1 \text{ hasStorageDevice})$. The latter concept can be safely conjoined to `Computer`—making the inclusion unnecessary—but cannot replace it since $(\geq 1 \text{ hasStorageDevice})$ is not a sufficient condition for `Computer`. Instead, `Computer` $\sqcap (\geq 2 \text{ hasCPU})$ replaces `Server` since it is a necessary and sufficient condition for it. The disjoint assertion generates `Intel` $\sqcap \neg$`AMD` as the range for $\forall$`hasCPU.`. Once this rewriting has been carried over all concepts, the TBox can be safely ignored when computing subsumption (and satisfiability). In general, this unfolding may lead to an exponential blow-up of the TBox, making the entire computation (unfolding+subsumption) take exponential time (and space) in the size of the initial concepts and TBox. Yet exponential-time computation for subsumption is likely to be unavoidable, since even without rewriting, taking the TBox into account makes subsumption NP-hard (Nebel, 1990).

The normal form of concepts can take the TBox embedding into account (see Appendix A.2). In this case, the component $C_{names}$ of a Classic concept $C$ contains concept names $C_{names+}$ and negations of concept names $C_{names\neg}$. In the following, we denote the CNF of a concept $C$ w.r.t. a simple TBox $\mathcal{T}$ as $CNF(C, \mathcal{T})$. Again, in general, the size of $CNF(C, \mathcal{T})$ may be exponential w.r.t. the size of $C$ and $\mathcal{T}$. However, when $\mathcal{T}$ is *fixed*, $CNF(C, \mathcal{T})$ has polynomial-size w.r.t. the size of $C$ *i.e.*, the exponential increase comes only from the TBox unfolding. In fact, if $k$ is the maximum size of an unfolded concept name (a constant if $\mathcal{T}$ is fixed), the size of $CNF(C, \mathcal{T})$ can be at most $k$ times the size of $C$. We use this argument later in the paper, to decouple the complexity analysis of our reasoning methods for matchmaking from the complexity raised by the TBox.

To ease presentation of what follows in the next Sections, we adopt a simple reference ontology, pictured in Figure 1, which is used throughout the paper. To keep the representation within $\mathcal{ALN}$, we modeled memory quantities with number restriction, *e.g.*, 20GB as $(\geq\ 20000\ \mathtt{mb})$. For reasoners specialized for $\mathcal{ALN}$, this is not a problem, since a number $n$ is never expanded as $n$ fillers (Borgida & Patel-Schneider, 1994; Donini et al., 1997b). For more expressive DLs, Concrete Domains (Lutz, 1999) should be employed to represent such quantities.

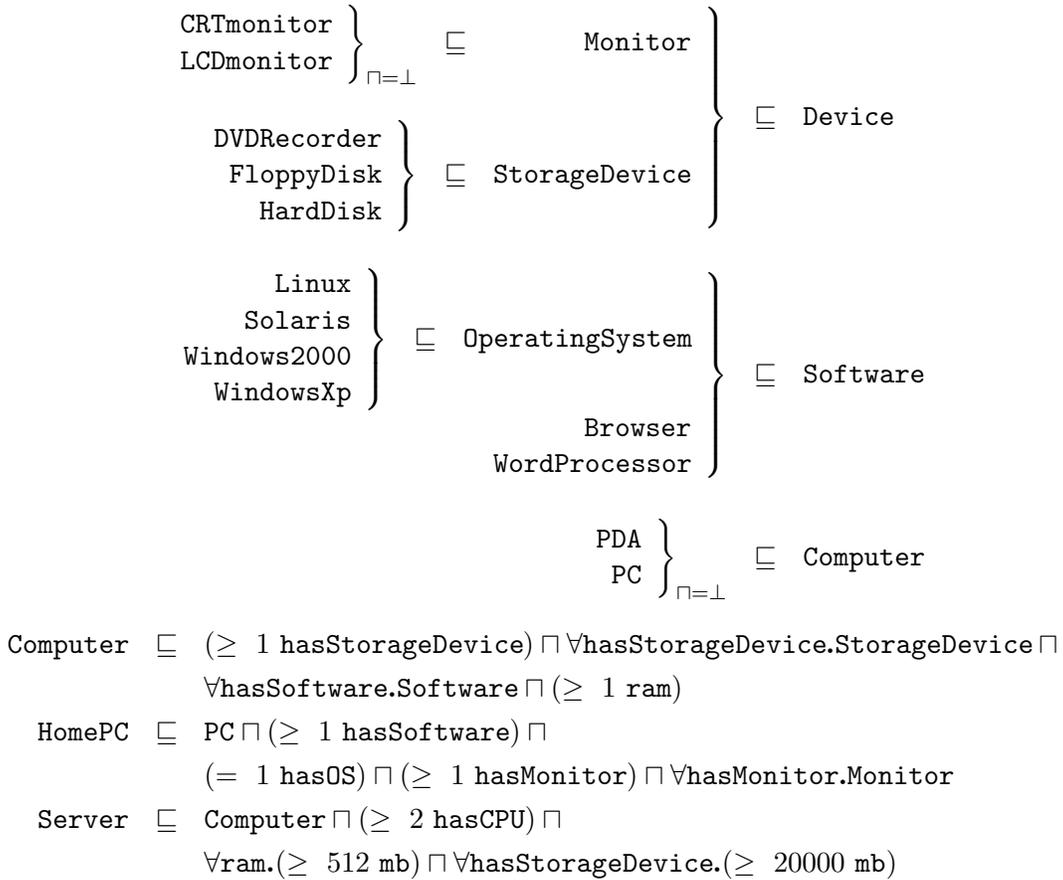

Figure 1: Reference Ontology used for examples

## 4. Semantic Matchmaking Using Description Logics

Matchmaking is a widely used term in a variety of frameworks, comprising several—quite different—approaches. We begin this Section trying to provide a generic and sound definition of matchmaking.

> **Matchmaking** is an information retrieval task whereby queries (a.k.a. demands) and resources (a.k.a. supplies) are expressed using semi-structured data in the form of advertisements, and task results are ordered (ranked) lists of those resources best fulfilling the query.

This simple definition implies that—differently from classical unstructured-text Information Retrieval systems—some structure in the advertisements is expected in a matchmaking system, and matchmaking does not consider a fixed database-oriented relational structure. Furthermore, usually database systems provide answers to queries that do not include a relevance ranking, which should be instead considered in a matchmaking process.

> **Semantic matchmaking** is a matchmaking task whereby queries and resources advertisements are expressed with reference to a shared specification of a conceptualization for the knowledge domain at hand, *i.e.*, an ontology.

From now on, we concentrate on semantic matchmaking in marketplaces, adopting specific terminology, to ease presentation of the approach. Nevertheless our approach applies to generic matchmaking of semantically annotated resources.

We note that all definitions in this Section apply to every DL that can be used to describe a marketplace (supplies, demands, background knowledge). We denote by $\mathcal{L}$ such a generic DL. We suppose that a common ontology for supplies and demands is established, as a TBox $\mathcal{T}$ in $\mathcal{L}$. Now a match between a supply and a demand could be evaluated according to $\mathcal{T}$.

First of all, we remark that a logic-based representation of supplies and demands calls for generally *Open-world descriptions*, that is, the absence of a characteristic in the description of a supply or demand should not be interpreted as a constraint of absence. Instead, it should be considered as a characteristic that could be either refined later, or left open if it is irrelevant for a user. Note that by "generally open" we mean that some specific characteristic might be declared to be closed. However, such a closure should be made piecewise, using some known declarative tool devised in Knowledge Representation for non-monotonic reasoning, such as Defaults in DLs (Baader & Hollunder, 1992), Autoepistemic DLs (Donini, Nardi, & Rosati, 1997a), Circumscription in DLs (Bonatti, Lutz, & Wolter, 2006) etc.

An analysis of recent literature allows to categorize the semantic matchmaking process between a supply *Sup* and a demand *Dem* w.r.t. a TBox $\mathcal{T}$ in five distinct classes:

- **exact match**: $Sup \equiv_\mathcal{T} Dem$, *i.e.*, $Sup \sqsubseteq_\mathcal{T} Dem$ and $Dem \sqsubseteq_\mathcal{T} Sup$, which amounts to a perfect match, regardless—in a semantic based environment—of syntactic differences, *i.e.*, *Sup* and *Dem* are equivalent concepts (Di Sciascio et al., 2001; Gonzales-Castillo et al., 2001).

- **full match**: $Sup \sqsubseteq_\mathcal{T} Dem$, which amounts to the demand being completely fulfilled by the available supply, *i.e.*, *Sup* has at least all features required by *Dem*, but not necessarily vice versa, being the matchmaking process not symmetric (Di Noia et al., 2003c); this kind of match is also named *subsume* match by Li and Horrocks (2003).

- **plug-in match**: $Dem \sqsubseteq_\mathcal{T} Sup$; it corresponds to demand *Dem* being sub-concept of supply *Sup*, *i.e.*, *Dem* is more specific than *Sup* (Sycara et al., 2002; Li & Horrocks, 2003).

- **potential match**: $Dem \sqcap Sup \not\sqsubseteq_\mathcal{T} \bot$, which corresponds to supply and demand having something in common and no conflicting characteristics (Di Noia et al., 2003c). This relation is also named *intersection-satisfiable* by Li and Horrocks (2003).

- **partial match**: $Dem \sqcap Sup \sqsubseteq_{\mathcal{T}} \bot$, which amounts to the presence of conflict between the demand and the available supply (Di Noia et al., 2003c). This relation is also named *disjoint* by Li and Horrocks (2003)[5].

We stress that demands could be classified in the same way w.r.t. a given supply, when it's the supplier's turn to look into the marketplace to find potential buyers. Hence, in the rest of the paper we use the term *offer*—denoted by the symbol $D$—to mean either a supply *Sup* or a demand *Dem*, and the term *counteroffer*—denoted by $C$—to mean, respectively, the demand *Dem* or the supply *Sup* that could match $D$.

Such a classification is still a coarse one, relying directly on known logical relations between formulae. In fact, the result of matchmaking should be a *rank* of counteroffers, according to some criteria—possibly explicit—so that a user trusting the system would know whom to contact first, and in case of failure, whom next, and so on. Such a ranking process should satisfy some criteria that a Knowledge Representation approach suggests.

We formulate ranking requirements by referring to properties of penalty functions.

**Definition 1** *Given a DL $\mathcal{L}$, two concepts $C, D \in \mathcal{L}$, and a TBox $\mathcal{T}$ in $\mathcal{L}$, a penalty function is a three-arguments function $p(C, D, \mathcal{T})$, that returns a null or positive integer.*

We use penalty functions to rank counteroffers $C$ for a given demand (or supply) $D$ w.r.t. a TBox $\mathcal{T}$. Intuitively, for two given counteroffers $C_1, C_2$ in the marketplace, if $p(C_1, D, \mathcal{T}) < p(C_2, D, \mathcal{T})$ then the issuer of offer $D$ should rank $C_1$ better than $C_2$ when deciding whom to contact first. Clearly, a 0-penalty should be ranked best, and counteroffers with the same penalties should be ranked breaking ties. The first property we recall is Non-symmetric evaluation of proposals.

**Definition 2** *A penalty function $p(\cdot, \cdot, \cdot)$ is non-symmetric if there exist concepts $C, D$ and a TBox $\mathcal{T}$ such that $p(C, D, \mathcal{T}) \neq p(D, C, \mathcal{T})$.*

This property is evident when all constraints of $D$ are fulfilled by $C$ but not vice versa. Hence, $C$ should be among the top-ranked counteroffers in the list of potential partners of $D$, while $D$ should *not* necessarily appear at the top in the list of potential partners of $C$. So, a penalty function $p(\cdot, \cdot, \cdot)$ should not be expected to be a metric distance function.

Secondly, if logic is used to give some meaning to descriptions of supplies and demands, then proposals with the same meaning should be equally penalized, independently of their syntactic descriptions.

**Definition 3** *A penalty function $p(\cdot, \cdot, \cdot)$ is syntax independent if for every triple of concepts $C_1, C_2, D$, and TBox $\mathcal{T}$, when $\mathcal{T} \models C_1 \equiv C_2$ then $p(C_1, D, \mathcal{T}) = p(C_2, D, \mathcal{T})$, and the same holds also for the second argument , i.e., $p(D, C_1, \mathcal{T}) = p(D, C_2, \mathcal{T})$*

---

5. We note that preferring the term "partial match" instead of "disjoint", we stress that the match may still be *recoverable*, while disjoint is usually meant as a *hopeless* situation. Moreover, "disjoint" and "intersection satisfiable" refer to the set-theoretic semantics of concepts in Description Logics, which is quite hidden and far from the original problems of matchmaking. In a word, they are technology-oriented and not problem-oriented. For instance, if one used Propositional Logic, or Three-valued Logic for modeling matchmaking, those terms would make no sense.

Clearly, when the logic admits a normal form of expressions—as CNF or DNF for propositional logic, or the normal form of concepts for DLs defined in the previous Section—using such a normal form in the computation of $p(\cdot, \cdot, \cdot)$ ensures by itself syntax independence.

Penalties should enjoy some desirable properties w.r.t. subsumption. For reasons explained below, we divide penalty functions for ranking potential matches from those for ranking partial (conflicting) matches.

**Definition 4** *A penalty function for potential matches is* monotonic over subsumption *whenever for every issued offer $D$, for every pair of counteroffers $C_1$ and $C_2$, and TBox $\mathcal{T}$, if $C_1$ and $C_2$ are both potential matches for $D$ w.r.t. $\mathcal{T}$, and $(C_1 \sqsubseteq_{\mathcal{T}} C_2)$, then $p(C_1, D, \mathcal{T}) \leq p(C_2, D, \mathcal{T})$*

Intuitively, the above definition could be read as of: if $C_1 \sqsubseteq_{\mathcal{T}} C_2$ then $C_1$ should be penalized (and then ranked) either the same, or better than $C_2$. In a phrase, *A ranking of potential matches is monotonic over subsumption if more specific means better.* A dual property could be stated for the second argument: if $D_1 \sqsubseteq_{\mathcal{T}} D_2$ then a counteroffer $C$ is less likely to fulfill all characteristics required by $D_1$ than $D_2$. However, since our scenario is: "given an issuer of a proposal $D$ looking for a match in the marketplace, rank all possible counteroffers $C_1, C_2, \ldots$, from the best one to the worst", we do not deal here with this duality between first and second argument of $p(\cdot, \cdot, \cdot)$.

When turning to partial matches, in which some properties are already in conflict between supply and demand, the picture reverses. Now, adding another characteristic to an unsatisfactory proposal may only worsen this ranking (when another characteristic is violated) or keep it the same (when the new characteristic is not in conflict). Note that this ranking should be kept different from the ranking for potential matches. After all, accepting to discard one or more characteristics that we required is much worse than deciding which proposal try first among some potential ones.

**Definition 5** *A penalty function for partial matches is* antimonotonic over subsumption *whenever for every issued offer $D$, for every pair of counteroffers $C_1$ and $C_2$, and TBox $\mathcal{T}$, if $C_1$ and $C_2$ are both partial matches for $D$ w.r.t. $\mathcal{T}$, and $(C_1 \sqsubseteq_{\mathcal{T}} C_2)$, then $p(C_1, D, \mathcal{T}) \geq p(C_2, D, \mathcal{T})$*

Intuitively, if $C_1 \sqsubseteq_{\mathcal{T}} C_2$ then $C_1$ should be penalized (and then ranked) either the same, or worse than $C_2$. In other words, *A ranking of partial matches is antimonotonic over subsumption if more specific means worse.* The same property should hold also for the second argument, since concept conjunction is commutative.

When we need to distinguish between a penalty function for potential matches and one for partial matches, we put a subscript $\sqsubseteq$ in the former (as $p_{\sqsubseteq}$) and a subscript $\bot$ for the latter (as in $q_{\bot}$).

Clearly, the above requirements are very general, and leave ample room for the definition of penalty functions. A more subtle requirement would be that penalties should not change when irrelevant details are added, *e.g.*, if a second-hand computer is requested in a demand *Dem*, with no specification for the brand of the CPU, then a supply *Sup* should be penalized the same as another offer $Sup \sqcap \forall\texttt{hasCPU.Intel}$. However, instead of delving into irrelevance and other logic-related issues directly from penalties, we now borrow well-known logical reasoning frameworks in propositional knowledge representation. Such a detour will give us a sound and declarative way of defining penalties, dealing with irrelevance as a byproduct, and more generally bring well-studied non-standard reasoning techniques into matchmaking.

## 5. Concept Abduction

Abduction (Peirce, 1955) is a well known form of commonsense reasoning, usually aimed at finding an explanation for some given symptoms or manifestations. Here we introduce Concept Abduction in DLs, showing how it can model potential matchmaking in a DL setting. Following the notation proposed by Eiter and Gottlob (1995), we recall that a Propositional Abduction Problem is a triple $\langle H, M, T \rangle$ where $H$ (Hypotheses) and $M$ (Manifestations) are sets of literals, and $T$ (Theory) is a set of formulae. A solution for $\langle H, M, T \rangle$ is an Explanation $E \subseteq H$ such that $T \cup E$ is consistent, and $T \cup E \models M$. We adapt this framework to DLs as follows.

**Definition 6** *Let $\mathcal{L}$ be a DL, $C$, $D$, be two concepts in $\mathcal{L}$, and $\mathcal{T}$ be a set of axioms in $\mathcal{L}$, where both $C$ and $D$ are satisfiable in $\mathcal{T}$. The* Concept Abduction Problem *(CAP) for a given $\langle \mathcal{L}, C, D, \mathcal{T} \rangle$, is finding, if possible, a concept $H \in \mathcal{L}$ such that $C \sqcap H \not\sqsubseteq_{\mathcal{T}} \bot$, and $C \sqcap H \sqsubseteq_{\mathcal{T}} D$.*

We use $\mathcal{P}$ as a symbol for a generic CAP, and we denote with $SOL(\mathcal{P})$ the set of all solutions to a CAP $\mathcal{P}$. Observe that in the definition, we limit the inputs of a CAP to satisfiable concepts $C$ and $D$, since $C$ unsatisfiable implies that the CAP has no solution at all, while $D$ unsatisfiable leads to counterintuitive results (*e.g.*, $\neg C$ would be a solution in that case). As Propositional Abduction extends implication, Concept Abduction extends concept subsumption. But differently from propositional abduction, we do not make any distinction between manifestations and hypotheses, which is usual when abduction is used for diagnosis. In fact, when making hypotheses about *e.g.*, properties of goods in e-marketplaces, there is no point in making such a distinction. This uniformity implies that there is always the trivial solution $D$ to a non-trivial CAP $\langle \mathcal{L}, C, D, \mathcal{T} \rangle$, as stated more formally as follows.

**Proposition 1** *Let $\mathcal{L}$ be a DL, let $C, D$ be concepts in $\mathcal{L}$, and $\mathcal{T}$ an $\mathcal{L}$-TBox. Then $C \sqcap D \not\sqsubseteq_{\mathcal{T}} \bot$ if and only if $D \in SOL(\langle \mathcal{L}, C, D, \mathcal{T} \rangle)$.*

*Proof.* If $C \sqcap D$ is satisfiable in $\mathcal{T}$, then $D$ fulfills both requirements of Def. 6, the first one by hypothesis and the second one because $C \sqcap D \sqsubseteq_{\mathcal{T}} D$ is a tautology. On the other hand, if $D \in SOL(\langle \mathcal{L}, C, D, \mathcal{T} \rangle)$ then $C \sqcap D \not\sqsubseteq_{\mathcal{T}} \bot$ by definition. □

A simple interpretation of this property in our application domain, *i.e.*, matchmaking, is that if we hypothesize for the counteroffer $C$ exactly all specifications in $D$, then the counteroffer trivially meets given specifications—if it was compatible anyway. However, not all solutions to a CAP are equivalent when using Concept Abduction for matchmaking. To make a simple example, suppose that already $C \sqsubseteq_{\mathcal{T}} D$. Then, both $H_1 = D$ and $H_2 = \top$ (among others) are solutions of $\langle \mathcal{L}, C, D, \mathcal{T} \rangle$. Yet, the solution $H_2 = \top$ tells the issuer of $D$ that $C$ already meets all of $D$'s specifications, while the solution $H_1 = D$ is the least informative solution from this point of view. Hence, if we want to use abduction to highlight most promising counteroffers, "minimal" hypotheses must be defined.

**Definition 7** *Let* $\mathcal{P} = \langle \mathcal{L}, C, D, \mathcal{T} \rangle$ *be a CAP. The set* $SOL_{\sqsubseteq}(\mathcal{P})$ *is the subset of* $SOL(\mathcal{P})$ *whose concepts are maximal under* $\sqsubseteq_{\mathcal{T}}$. *The set* $SOL_{\leq}(\mathcal{P})$ *is the subset of* $SOL(\mathcal{P})$ *whose concepts have minimum length.*

Clearly, being maximal w.r.t. $\sqsubseteq_{\mathcal{T}}$ is still a minimality criterion, since it means that no unnecessary hypothesis is assumed. It can be proved that the two measures are incomparable.

**Proposition 2** *There exists a CAP* $\mathcal{P}$ *such that the two sets* $SOL_{\sqsubseteq}(\mathcal{P})$ *and* $SOL_{\leq}(\mathcal{P})$ *are incomparable.*

*Proof.* It is sufficient to consider $D = A_1 \sqcap A_2 \sqcap A_3$, $C = A_1$, and $\mathcal{T} = \{B \sqsubseteq A_2 \sqcap A_3\}$. The logic is even propositional. Then $A_2 \sqcap A_3 \in SOL_{\sqsubseteq}(\langle \mathcal{L}, C, D, \mathcal{T} \rangle)$, $B \in SOL_{\leq}(\langle \mathcal{L}, C, D, \mathcal{T} \rangle)$, and neither solution is in the other set. □

The proof highlights that, although $\leq$-minimality could be preferable for conciseness, it is heavily dependent on $\mathcal{T}$. In fact, for every concept $H \in SOL(\mathcal{P})$, it is sufficient to add the axiom $A \equiv H$ to get a $\leq$-minimal solution $A$. On the other hand, also $\sqsubseteq_{\mathcal{T}}$-maximality has some drawbacks: if concept disjunction $\sqcup$ is present in $\mathcal{L}$, then there is a single $\sqsubseteq_{\mathcal{T}}$-maximal solution of $\mathcal{P}$, that is equivalent to the disjunction of all solutions in $SOL(\mathcal{P})$—not a very useful solution. Making an analogy with Abduction-based Diagnosis (Console, Dupre, & Torasso, 1991), we could say that the disjunction of all possible explanations is not a very informative explanation itself—although it is maximal w.r.t. implication. We note that finding a $\leq$-minimal solution is NP-hard for a TBox of depth 1, by a simple reduction from Set Covering (Colucci, Di Noia, Di Sciascio, Donini, & Mongiello, 2004).

**Remark 1** It is interesting to analyze whether concept minimal-rewriting techniques—as defined by Baader, Küsters, and Molitor (2000)—could be employed for computing some minimal concept abduction, trying to rewrite $C \sqcap D$. The answer is definitely negative for minimal length abduction: the length-minimal solution $B$ in the proof of Proposition 2 could not be obtained by rewriting $C \sqcap D = A_1 \sqcap A_1 \sqcap A_2 \sqcap A_3$. In fact, $A_1 \sqcap B$ is not an equivalent rewriting of the former concept. Regarding $\sqsubseteq_{\mathcal{T}}$-maximality the answer is more indirect. In fact, present rewriting techniques do not keep a subconcept fixed in the rewriting process. So consider a CAP in which $D = A_1$, $C = A_2$, and $\mathcal{T} = \{B \equiv A_1 \sqcap A_2\}$. The only equivalent minimal rewriting of $C \sqcap D$ is then $B$, in which a solution cannot be identified since $B$ cannot be separated into a concept $C$—the original one—and a concept $H$ that is a solution of the CAP. It is open whether future extensions of rewriting might keep a concept fixed, and cope with this problem.

A third minimality criterion is possible for DLs which admit CNF, as for $\mathcal{L} = \mathcal{ALN}$.

**Definition 8** *Let* $\mathcal{P} = \langle \mathcal{L}, C, D, \mathcal{T} \rangle$ *be a CAP in which* $\mathcal{L}$ *admits CNF, and assume that concepts in* $SOL(\mathcal{P})$ *are in CNF. The set* $SOL_{\sqcap}(\mathcal{P})$ *is the subset of* $SOL(\mathcal{P})$ *whose concepts are minimal conjunctions, i.e., if* $C \in SOL_{\sqcap}(\mathcal{P})$ *then no sub-conjunction of* $C$ *(at any level of nesting) is in* $SOL(\mathcal{P})$. *We call such solutions* irreducible.

It turns out that $\sqcap$-minimality includes both $\sqsubseteq_{\mathcal{T}}$-maximality and $\leq$-minimality.

**Proposition 3** *For every CAP $\mathcal{P}$ in which $\mathcal{L}$ admits a CNF, both $SOL_{\sqsubseteq}(\mathcal{P})$ and $SOL_{\leq}(\mathcal{P})$ are included in $SOL_{\sqcap}(\mathcal{P})$.*

*Proof.*   By contraposition, if a concept $H$ is not $\sqcap$-minimal then there is another concept $H'$—a sub-conjunction of $H$—which is an $\sqcap$-minimal solution. But $|H'| < |H|$, hence $H$ is not length-minimal. The same for $\sqsubseteq_{\mathcal{T}}$-maximality: since every sub-conjunction of a concept $H$ in CNF subsumes $H$, if $H$ is not $\sqcap$-minimal it is not $\sqsubseteq_{\mathcal{T}}$-maximal either.   $\square$

The proof of Proposition 2 can be modified to show that minimum-length abduced concepts are not unique: it is sufficient to add another axiom $B' \sqsubseteq A_2 \sqcap A_3$ to obtain another minimum-length solution $B'$. A less obvious result is that also subsumption-maximal solutions are not unique, at least in non-simple TBoxes: Let $\mathcal{P} = \langle \mathcal{L}, C, D, \mathcal{T} \rangle$ with $\mathcal{T} = \{A_2 \sqcap A_3 \sqsubseteq A_1\}$, $C = A_3, D = A_1$. Then both $A_1$ and $A_2$ are $\sqsubseteq_{\mathcal{T}}$-maximal solutions.

### 5.1 Irreducible Solutions in $\mathcal{ALN}$-simple TBoxes

We assume here that the TBox $\mathcal{T}$ of a CAP $\mathcal{P} = \langle \mathcal{L}, C, D, \mathcal{T} \rangle$ is always a simple one. Finding an irreducible solution is easier than finding a $\leq$-minimal or a $\sqsubseteq_{\mathcal{T}}$-maximal solution, since a greedy approach can be used to minimize the set of conjuncts in the solution. For example, starting from $C \sqcap D$, we could delete one redundant conjunct at a time (at any level of role quantification nesting) from $D$, using $|D|$ calls to a subsumption-check procedure. However, such an algorithm would be interesting only for theoretical purposes. Instead, we adapt a structural subsumption algorithm (Borgida & Patel-Schneider, 1994) that collects all concepts $H$ that should be conjoined to $C$ in order for $C \sqcap H$ to be subsumed by $D$. The algorithm operates on concepts in CNF. In the following algorithm, we abbreviate the fact that a concept $A$ appears as a conjunct of a concept $C$ with $A \in C$ (thus extending the meaning of $\in$ to conjunctions of concepts).

**Algorithm** *findIrred*($\mathcal{P}$);
    **input:** a CAP $\mathcal{P} = \langle \mathcal{L}, C, D, \mathcal{T} \rangle$, with $\mathcal{L} = \mathcal{ALN}$, simple $\mathcal{T}$, $C$ and $D$ in CNF w.r.t. $\mathcal{T}$
    **output:** concept $H \in SOL_{\sqcap}(\mathcal{P})$ (where $H = \top$ means that $C \sqsubseteq D$)
    **variables:** concept $H$
**begin**
    $H := \top$;
0.  **if** $D \sqcap C \sqsubseteq_{\mathcal{T}} \bot$
        **return** $\bot$;
1.  **for every** concept name $A$ in $D$
1.1     **if** $A \notin C$
        **then** $H := H \sqcap A$;
2.  **for every** concept $(\geq\ n\ R) \in D$
2.1     **such that** there is no concept $(\geq\ m\ R) \in C$ with $m \geq n$
        $H := H \sqcap (\geq\ n\ R)$;
3.  **for every** concept $(\leq\ n\ R) \in D$

3.1     **such that** there is no concept $(\leq\ m\ R) \in C$ with $m \leq n$

        $H := H \sqcap (\leq\ n\ R)$;

4. **for every** concept $\forall R.E \in D$

4.1     **if** there exists $\forall R.F \in C$

4.1.1        **then** $H := H \sqcap \forall R.findIrred(\langle \mathcal{ALN}, F, E, \mathcal{T} \rangle)$;

4.1.2        **else** $H := H \sqcap \forall R.E$;

/* now $H \in SOL(\mathcal{P})$, but it might be reducible */

5. **for every** concept $H_i \in H$

    **if** $H$ without $H_i$ is in $SOL(\mathcal{P})$

    **then** delete $H_i$ from $H$;

6. **return** $H$;

**end.**

**Theorem 1** *Given a CAP $\mathcal{P}$, if findIrred($\mathcal{P}$) returns the concept $H$, with $H \not\equiv \bot$, then $H$ is an irreducible solution of $\mathcal{P}$.*

*Proof.*   We first prove that before Step 5, the computed concept $H$ is in $SOL(\mathcal{P})$, that is, both $C \sqcap H \not\sqsubseteq_{\mathcal{T}} \bot$ and $C \sqcap H \sqsubseteq_{\mathcal{T}} D$ hold. In fact, observe that $CNF(D, \mathcal{T}) \sqsubseteq H$, since all conjuncts of $H$ come from some conjunct of $CNF(D, \mathcal{T})$. Hence, $D \sqsubseteq_{\mathcal{T}} H$ since $CNF(D, \mathcal{T})$ is equivalent to $D$ in the models of $\mathcal{T}$. Adding $C$ to both sides of the subsumption yields $C \sqcap D \sqsubseteq_{\mathcal{T}} C \sqcap H$, and since we assume that $C \sqcap D \not\sqsubseteq_{\mathcal{T}} \bot$, also $C \sqcap H \not\sqsubseteq_{\mathcal{T}} \bot$. This proves the first condition for $H \in SOL(\mathcal{P})$. Regarding the condition $C \sqcap H \sqsubseteq_{\mathcal{T}} D$, suppose it does not hold: then, at least one conjunct of $CNF(D, \mathcal{T})$ should not appear in $CNF(C \sqcap H, \mathcal{T})$. But this is not possible by construction, since $H$ contains every conjunct which is in $CNF(D, \mathcal{T})$ and not in $CNF(C, \mathcal{T})$. Therefore, we conclude that $H \in SOL(\mathcal{P})$. Once we proved that the $H$ computed before Step 5 is a solution of $\mathcal{P}$, we just note that Step 5 deletes enough conjuncts to make $H$ an irreducible solution. □

The first part of algorithm (before Step 5) easily follows well-known structural subsumption algorithms (Borgida & Patel-Schneider, 1994). Step 5 applies a greedy approach, hence the computed solution, although irreducible, might not be minimal.

We explain the need for the reducibility check in Step 5 with the help of the following example.

**Example 1** Let $\mathcal{T} = \{A_1 \sqsubseteq A_2, A_3 \sqsubseteq A_4\}$, and let $C = A_3$, $D = A_1 \sqcap A_4$. Then $\mathcal{L}$ is the propositional part of $\mathcal{AL}$. The normal form for $C$ is $C' = A_3 \sqcap A_4$, while $D' = A_1 \sqcap A_2 \sqcap A_4$. Then before Step 5 the algorithm computes $H = A_1 \sqcap A_2$, which must still be reduced to $A_1$. It is worth noticing that $H$ is already subsumption-maximal since $H \equiv_T A_1$. However, $\sqcap$-minimality is a syntactic property, which requires removal of redundant conjuncts.

As for complexity, we aim at proving that finding an irreducible solution is not more complex than subsumption in $\mathcal{ALN}$. A polynomial algorithm (w.r.t. the sizes of $C$, $D$ and $\mathcal{T}$) cannot be expected anyway, since subsumption in $\mathcal{AL}$ (the sublanguage of $\mathcal{ALN}$ without Number Restrictions) with a simple $\mathcal{T}$ is coNP-hard (Nebel, 1990; Calvanese, 1996). However, Nebel (1990) argues that the unfolding of the TBox is exponential in the depth of the hierarchy $\mathcal{T}$; if the depth of $\mathcal{T}$ grows as $O(\log|\mathcal{T}|)$ as the size of $\mathcal{T}$ increases—a "bushy but not deep" TBox—then its unfolding is polynomial, and so is the above algorithm.

More generally, suppose that $\mathcal{T}$ is *fixed*: this is not an unrealistic hypothesis for our marketplace application, since $\mathcal{T}$ represents the ontology of the domain, that we do not expect to vary while supplies and demands enter and exit the marketplace. In that case, we can analyze the complexity of *findIrred* considering only $C$ and $D$ for the size of the input of the problem.

**Theorem 2** *Let $\mathcal{P} = \langle \mathcal{L}, C, D, \mathcal{T} \rangle$ be a CAP, with $\mathcal{L} = \mathcal{ALN}$, and $\mathcal{T}$ a simple TBox. Then finding an irreducible solution to $\mathcal{P}$ is a problem solvable in time polynomial in the size of $C$ and $D$.*

We note that the problem of the exponential-size unfolding might be mitigated by Lazy Unfolding (Horrocks & Tobies, 2000). Using this technique, concept names in the TBox are unfolded only when needed.

## 5.2 Abduction-Based Ranking of Potential Matches

We define a penalty function $p_\sqsubseteq$ for potential matches based on the following intuition: the ranking of potential matches should depend on how many hypotheses have to be made on counteroffers in order to transform them into full matches.

**Definition 9** *Given a simple TBox $\mathcal{T}$ in $\mathcal{ALN}$, we define a penalty function for the potential match of a counteroffer $C$ given an offer $D$, where both $C$ and $D$ are concepts in $\mathcal{ALN}$, as follows:*

$$p_\sqsubseteq(C, D, \mathcal{T}) \doteq |findIrred(\langle \mathcal{ALN}, CNF(C, \mathcal{T}), CNF(D, \mathcal{T}), \emptyset \rangle)| \qquad (1)$$

Note that, when computing $p_\sqsubseteq$, a concept $H$ is actually computed by *findIrred* as an intermediate step. This makes it easy to devise an explanation facility, so that the actual obtained ranking can be immediately enriched with its logical explanation; thus improving users' trust and interaction with the matchmaking system.

We now prove that $p_\sqsubseteq$ is in accordance with properties higlighted in the previous Section. Since the computation of Formula (1) starts by putting concepts $C, D$ in normal form, we recall that the normal form of $C$ can be summarized as $C_{names} \sqcap C_\sharp \sqcap C_{all}$, and similarly for $D$. Without ambiguity, we use the three components also as sets of the conjoined concepts.

**Theorem 3** *The penalty function $p_\sqsubseteq$ is (i) non-symmetric, (ii) syntax independent, and (iii) monotonic over subsumption.*

*Proof.* (i) Non-symmetricity is easily proved by providing an example: $p_\sqsubseteq(A, \top, \emptyset) \neq p_\sqsubseteq(\top, A, \emptyset)$. In fact, *findIrred* $(\langle \mathcal{ALN}, A, \top, \emptyset \rangle)$ finds $H_1 = \top$ as a solution ($A \sqsubseteq \top$ without further hypothesis) while *findIrred* $(\langle \mathcal{ALN}, \top, A, \emptyset \rangle)$ finds $H_2 = A$. Recalling that $|\top| = 0$, while $|A| = 1$, we get the first claim.

(ii) Syntax independence follows from the fact that normal forms are used in Formula (1), and as already said normal forms are unique up to commutativity of conjunction.

(iii) Monotonicity over subsumption is proved by analyzing the conditions for subsumption in $\mathcal{ALN}$. A concept $C'$ is subsumed by a concept $C$ whenever all conditions below hold. For each condition, we analyze the changes in the behavior of *findIrred*, proving that the provided solution $H$ just adds other conjuncts. Recall that monotonicity over subsumption is applied only to potential matches, hence we assume that both $C$ and $C'$ are consistent with $D$. Since *findIrred* is recursive, the proof is also by induction on the quantification nesting (QN) of $C'$. For $C'$ having QN equal to 0, $C'$ can only be a conjunction of atomic concepts—names, negated names, number restrictions. Then the conditions for subsumption are the following:

- The first condition is that $C_{names+} \subseteq C'_{names+}$. Hence, in Step 1.1 of *findIrred*, the number of concept names that are added to $H'$—with respect to names added to $H$— can only decrease, and so $|H'| \leq |H|$ considering names. Regarding negated names, observe that they do not contribute to the solution of *findIrred*, since they come from a disjointness axiom and a positive name (that contributes).

- The second condition is that for every number restriction in $C_\sharp$, either the same number restriction appears in $C'_\sharp$, or it is strengthened (an at-least increases, an at-most decreases) in $C'_\sharp$. Hence, number restrictions added by Steps 2.1 and 3.1 to $H'$ can be either as many as those added to $H$, or less. Again, also considering number restrictions $|H'| \leq |H|$.

The above two cases prove the basis of the induction ($C'$ with QN equal to 0). Suppose now the claim holds for concepts $C'$ with QN $n$ or less, and let $C'$ have a QN of $n+1$. Clearly, in this case $C'$ has at least one universal role quantification—call it $\forall R.F'$. The condition for subsumption between $C'$ and $C$ is the following:

- Either for every universal role quantification $\forall R.F$ in $C$ over the same role $R$, it must hold $F' \sqsubseteq_{\mathcal{T}} F$, or there is no universal role quantification on $R$ in $C$. In the former case, observe that *findIrred* is recursively called[6] in Step 4.1.1 with arguments $F$, $E$, and $F'$, $E$; we call $I$ and $I'$, respectively, the solutions returned by *findIrred*. Observe that the QN of $F'$ is $n$ or less, hence by inductive hypothesis $|I'| \leq |I|$. Since Step 4.1.1 adds $\forall R.I'$ and $\forall R.I$ to $H'$ and $H$, again $|H'| \leq |H|$. If instead there is no universal role quantification on $R$ in $C$, Step 4.1.2 adds $\forall R.E$ to $H$. If also $C'$ does not contain any role quantification on $R$, then Step 4.1.2 adds $\forall R.E$ also to $H'$, then $H'$ cannot be longer than $H$ in this case. If a role quantification $\forall R.F'$ is in $C'$, then Step 4.1.1 makes a recursive call with arguments $F', E$. In this case, the solution returned $I'$ has length less than or equal to $|E|$, hence the length of $H'$ cannot be longer than the length of $H$ also in this case.

In summary, if $C' \sqsubseteq_{\mathcal{T}} C$ then in no case the length of $H'$ increases with respect to the length of $H$. This proves the monotonicity over subsumption of $p_{\sqsubseteq}$. □

Intuitively, we could say that monotonicity over subsumption for potential matches means "the more specific $C$ is, the lower its penalty, the better its ranking w.r.t. $D$". More

---

6. *findIrred* is called only once, because concepts in CNF have at most one universal role quantification over any role $R$.

precisely—but less intuitively—we should say that "the rank of $C$ w.r.t. $D$ cannot worsen when $C$ is made more specific". Hence, given an offer $D$, a TBox $\mathcal{T}$, a sequence of increasingly specific counteroffers $C_1 \sqsupseteq_{\mathcal{T}} C_2 \sqsupseteq_{\mathcal{T}} C_3 \sqsupseteq_{\mathcal{T}} \cdots$ are assigned to a sequence of non-increasing penalties $p_{\sqsubseteq}(C_1, D, \mathcal{T}) \geq p_{\sqsubseteq}(C_2, D, \mathcal{T}) \geq p_{\sqsubseteq}(C_3, D, \mathcal{T}) \geq \ldots$ We now prove that such sequences are well-founded, with bottom element zero, reached in case of subsumption.

**Proposition 4** $p_{\sqsubseteq}(C, D, \mathcal{T}) = 0$ *if and only if* $C \sqsubseteq_{\mathcal{T}} D$.

*Proof.*　Recall from Section 3.1 that $\top$ and $\bot$ are the only concepts of length zero, and *findIrred* returns $\bot$ if and only if $C$ and $D$ are not in a potential match (Step 0 in *findIrred*). Hence, $p_{\sqsubseteq}(C, D, \mathcal{T}) = 0$ if and only if the concept whose length is computed in Formula (1) is $\top$. By construction of *findIrred*, $\top$ is returned by the call
*findIrred*$(\langle \mathcal{ALN}, CNF(C, \mathcal{T}), CNF(D, \mathcal{T}), \emptyset \rangle)$ if and only if $CNF(C, \mathcal{T}) \sqsubseteq CNF(D, \mathcal{T})$, which holds (see Borgida & Patel-Schneider, 1994) if and only if $C \sqsubseteq_{\mathcal{T}} D$.　　　□

Moreover, we could also prove that adding to $C$ details that are irrelevant for $D$ leaves the penalty unaffected, while adding to $C$ details that are relevant for $D$ lowers $C$'s penalty.

Note also that in Formula (1) we take $\mathcal{T}$ into account in the normal form of $C, D$, but then we forget it—we use an empty TBox—when calling *findIrred*. We explain such a choice with the aid of an example.

**Example 2** Given $\mathcal{T} = \{A \sqsubseteq A_1 \sqcap A_2\}$, let $D = A$ be a Demand with the two following supplies: $C_1 = A_2$, $C_2 = \top$. Observe that $CNF(D, \mathcal{T}) = A \sqcap A_1 \sqcap A_2$, $CNF(C_1, \mathcal{T}) = A_2$, $CNF(C_2, \mathcal{T}) = \top$. If we used the following formula to compute the penalty

$$p'(C, D, \mathcal{T}) \doteq |findIrred(\langle \mathcal{ALN}, C, D, \emptyset \rangle)| \qquad (2)$$

and ran the algorithm *findIrred*$(\langle \mathcal{ALN}, C_1, D, \mathcal{T} \rangle)$ and *findIrred*$(\langle \mathcal{ALN}, C_2, D, \mathcal{T} \rangle)$, before Step 5 we would get, respectively,

$$
\begin{aligned}
H_1 &= A_1 \sqcap A \\
H_2 &= A_1 \sqcap A_2 \sqcap A
\end{aligned}
$$

and after Step 5 *findIrred* would return $H'_1 = H'_2 = A$, hence $C_1$ and $C_2$ would receive the same penalty. However, we argue that $C_1$ is closer to $D$ than $C_2$ is, because it contains a characteristic ($A_2$) implicitly required by $D$, while $C_2$ does not. If instead we call *findIrred*$(\langle \mathcal{ALN}, CNF(C_1, \mathcal{T}), CNF(D, \mathcal{T}), \emptyset \rangle)$ and
*findIrred*$(\langle \mathcal{ALN}, CNF(C_2, \mathcal{T}), CNF(D, \mathcal{T}), \emptyset \rangle)$, we get the solutions $H_1$ and $H_2$ above—and Step 5 does not delete any conjunct, since $\mathcal{T} = \emptyset$. Therefore, $C_1$ gets penalty 2, while $C_2$ gets penalty 3, highlighting what is more specified in $C_1$ w.r.t. $C_2$.

More generally, we can say that the reducibility step (Step 5 in *findIrred*) flattens a solution to its most specific conjuncts, leaving to the TBox the implicit representation of other characteristics, both the ones already present in the supply and those not present. Therefore, making an empirical decision, we consider the TBox in the normal form of $C$ and $D$, but we exclude it from further reductions in Step 5 of *findIrred*.

**Remark 2** Although the definition of Concept Abduction could appear similar to Concept Difference, it is not so. We note that generically speaking, the name "Concept Abduction" appeals to logic, while "Concept Difference" appeals to algebra (although Difference has multiple solutions when $\mathcal{L}$ includes universal role quantification). More precisely, we recall (Teege, 1994) that difference is defined as: $C - D = \max_{\sqsubseteq} \{E \in L : (E \sqcap D) \equiv C\}$ provided that $C \sqsubseteq D$. A more specialized definition of difference (Brandt, Küsters, & Turhan, 2002) refers only to DLs $\mathcal{ALC}$ and $\mathcal{ALE}$. It is defined as: $C - D = \min_{\preceq} \{E \in L : (E \sqcap D) \equiv (C \sqcap D)\}$—where $C, E \in \mathcal{ALC}$, $D \in \mathcal{ALE}$, and minimality is w.r.t. a preorder $\preceq$ on a specific normal form which extends CNF to $\mathcal{ALC}$. No TBox is taken into account.

Instead, the solution of a CAP $\langle \mathcal{L}, C, D, \mathcal{T} \rangle$ does not require that $C \sqsubseteq_{\mathcal{T}} D$, but only that $C \sqcap D \not\sqsubseteq_{\mathcal{T}} \bot$. In general, when $D \sqsubseteq_{\mathcal{T}} C$ if we let $H = D - C$ in a CAP $\mathcal{P} = \langle \mathcal{L}, C, D, \mathcal{T} \rangle$ we get those solutions for which $C \sqcap H \equiv D$—which obviously are not all solutions to $\mathcal{P}$. Hence $D - C \subseteq SOL(\mathcal{P})$, but not vice versa (see the proof of Proposition 2 for an example). When $C \not\sqsubseteq_{\mathcal{T}} D$ this comparison is not even possible, since $D - C$ is undefined. However, in a generic setting, *e.g.*, in an e-commerce scenario, subsumption between demand and supply is quite uncommon; most of offers are such that neither subsumes the other. Because of this greater generality, for our specific application to matchmaking, Concept Abduction seems more suited than Concept Difference to make a basis for a penalty function.

## 6. Concept Contraction

If $D \sqcap C$ is unsatisfiable in $\mathcal{T}$, but the demander accepts to retract some of $D$'s constraints, partially matching supplies may be reconsidered. However, other logic-based approaches to matchmaking by Trastour et al. (2002), Sycara et al. (2002), Li and Horrocks (2003) usually exclude the case in which the concept expressing a demand is inconsistent with the concept expressing a supply, assuming that all requirements are strict ones. In contrast, we believe that inconsistent matches can still be useful, especially in e-marketplaces. In fact, partial (a.k.a. disjoint) matches can be the basis for a negotiation process, allowing a user to specify negotiable requirements—some of which could be bargained in favor of other. Such a negotiation process can be carried out in various ways adopting approaches to matchmaking not based on logic (*e.g.*, Ströbel & Stolze, 2002), but also, as shown in practice by Colucci et al. (2005), using Belief Revision. In fact, the logical formalization of conflicting matches, aimed at finding still "interesting" inconsistent matches without having to revert to text-based or hybrid approaches, can be obtained exploiting definitions typical of Belief Revision. In accordance with Gärdenfors (1988) formalization, revision of a knowledge base $\mathcal{K}$ with a new piece of knowledge $A$ is a *contraction* operation, which results in a new knowledge base $\mathcal{K}_A^-$ such that $\mathcal{K}_A^- \not\models \neg A$, followed by the addition of $A$ to $\mathcal{K}_A^-$—usually modeled by conjunction. We call *Concept Contraction* our adaptation of Belief Revision to DLs.

Starting with $C \sqcap D$ unsatisfiable in a TBox $\mathcal{T}$, we model with Concept Contraction how, retracting requirements in $C$, we may still obtain a concept $K$ (for Keep) such that $K \sqcap D$ is satisfiable in $\mathcal{T}$. Clearly, a user is interested in what he/she must negotiate on to start the transaction—a concept $G$ (for Give up) such that $C \equiv G \sqcap K$.

For instance, with reference to the ontology in Figure 1, if a user demands *Dem* and a supplier offers *Sup*, where *Dem* and *Sup* are described as follows:

$$Dem = \texttt{HomePC} \sqcap \forall\texttt{hasMonitor.LCDmonitor}$$
$$Sup = \texttt{HomePC} \sqcap \forall\texttt{hasMonitor.CRTmonitor}$$

it is possible to check that $Sup \sqcap Dem$ is unsatisfiable. This is a *partial match*. Yet, in this case, if the demander gives up the concept $G = \forall\texttt{hasMonitor.LCDmonitor}$ and keeps the concept $K = \texttt{HomePC}$, $K \sqcap Sup$ is satisfiable, hence $K$ now potentially matches *Sup*.

More formally we model a Concept Contraction problem as follows.

**Definition 10 (Concept Contraction)** *Let $\mathcal{L}$ be a DL, $C$, $D$, be two concepts in $\mathcal{L}$, and $\mathcal{T}$ be a set of axioms in $\mathcal{L}$, where both $C$ and $D$ are satisfiable in $\mathcal{T}$. A* Concept Contraction Problem *(CCP), denoted as $\langle \mathcal{L}, C, D, \mathcal{T} \rangle$, is finding a pair of concepts $\langle G, K \rangle \in \mathcal{L} \times \mathcal{L}$ such that $\mathcal{T} \models C \equiv G \sqcap K$, and $K \sqcap D$ is satisfiable in $\mathcal{T}$. We call $K$ a* contraction *of $C$ according to $D$ and $\mathcal{T}$.*

We use $\mathcal{Q}$ as a symbol for a CCP, and we denote with $SOLCCP(\mathcal{Q})$ the set of all solutions to a CCP $\mathcal{Q}$. Observe that as for concept abduction, we rule out cases where either $C$ or $D$ are unsatisfiable, as they correspond to counterintuitive situations. We note that there is always the trivial solution $\langle G, K \rangle = \langle C, \top \rangle$ to a CCP. This solution corresponds to the most drastic contraction, that gives up everything of $C$. On the other hand, when $C \sqcap D$ is satisfiable in $\mathcal{T}$, the "best" possible solution is $\langle \top, C \rangle$, that is, give up nothing.

As Concept Abduction extends Subsumption, Concept Contraction extends satisfiability—in particular, satisfiability of a conjunction $C \sqcap D$. Hence, results about the complexity of deciding Satisfiability of a given concept carry over to Contraction.

**Proposition 5** *Let $\mathcal{L}$ be a DL containing $\mathcal{AL}$, and let Concept Satisfiability w.r.t. a TBox in $\mathcal{L}$ be a problem $\mathcal{C}$-hard for a complexity class $\mathcal{C}$. Then deciding whether a given pair of concepts $\langle G, K \rangle$ is a solution of a CCP $\mathcal{Q} = \langle \mathcal{L}, C, D, \mathcal{T} \rangle$ is $\mathcal{C}$-hard.*

*Proof.* A concept $E \in \mathcal{L}$ is satisfiable w.r.t. a TBox $\mathcal{T}$ if and only if the CCP $\langle \mathcal{L}, C, D, \mathcal{T} \rangle$ has the solution $\langle \top, C \rangle$, where $C = \forall R.E$ and $D = \exists R.\top$. Then, $\mathcal{L}$ should contain at least universal role quantification (to express $\forall R.E$), unqualified existential role quantification (to express $\exists R.\top$), conjunction (to express that $C \equiv G \sqcap K$) and at least the unsatisfiable concept $\bot$ (otherwise every concept is satisfiable, and the problem trivializes). The minimal, known DL containing all such constructs is the DL $\mathcal{AL}$. □

This gives a lower bound on the complexity of Concept Contraction, for all DLs that include $\mathcal{AL}$. For DLs not including $\mathcal{AL}$, note that if the proof showing $\mathcal{C}$-hardness of satisfiability involves a concept with a topmost $\sqcap$ symbol, the same proof could be adapted for Concept Contraction.

Obviously, a user in a marketplace is likely to be willing to give up as few things as possible, so some minimality in the contraction $G$ must be defined. We skip for conciseness the definitions of a minimal-length contraction and subsumption-maximal contraction, and define straightforwardly conjunction-minimal contraction for DLs that admit a normal form made up of conjunctions.

**Definition 11** *Let $\mathcal{Q} = \langle \mathcal{L}, C, D, \mathcal{T} \rangle$ be a CCP in which $\mathcal{L}$ admits a CNF. The set $SOLCCP_{\sqcap}(\mathcal{Q})$ is the subset of $SOLCCP(\mathcal{Q})$ with the following property: if $\langle G, K \rangle \in SOLCCP_{\sqcap}(\mathcal{Q})$ then for no sub-conjunction $G'$ of $G$ it holds $\langle G', K \rangle \in SOLCCP(\mathcal{Q})$. We call such solutions irreducible.*

## 6.1 Number-Restriction Minimal Contractions

In what follows we focus on a specific class of irreducible solutions for a CCP $\langle \mathcal{ALN}, C, D, \mathcal{T} \rangle$ exposing interesting characteristics from a user-oriented point of view in a matchmaking scenario. Before defining such a class we explain the rationale behind its investigation using the following example.

**Example 3** Suppose we have the following situation:

$$
\begin{array}{llll}
\text{demand} & Dem & = & \texttt{HomePC} \sqcap \forall \texttt{hasMonitor.LCDmonitor} \\
\text{supply} & Sup & = & \texttt{Server} \sqcap \forall \texttt{hasMonitor.CRTmonitor}
\end{array}
$$

As $\mathcal{T} \models Dem \sqcap Sup \equiv \bot$ the demander can contract $Dem$ in order to regain the satisfiability with $Sup$. Two solutions for the CCP $\mathcal{Q} = \langle \mathcal{ALN}, Dem, Sup, \mathcal{T} \rangle$ are:

$$
\left\{
\begin{array}{lll}
G_{\geq} & = & \texttt{HomePC} \\
K_{\geq} & = & \texttt{PC} \sqcap (\geq 1 \ \texttt{hasSoftware}) \sqcap (= 1 \ \texttt{hasOS}) \\
& & \sqcap \ \forall \texttt{hasMonitor.LCDmonitor}
\end{array}
\right.
$$

$$
\left\{
\begin{array}{lll}
G_{\forall} & = & \forall \texttt{hasMonitor.LCDmonitor} \\
K_{\forall} & = & \texttt{HomePC}
\end{array}
\right.
$$

In $\langle G_{\geq}, K_{\geq} \rangle$ the demander should give up the specification on $\texttt{HomePC}$; in $\langle G_{\forall}, K_{\forall} \rangle$ the demander should give up only some specifications on the monitor type while keeping the rest.

Observe that both solutions are in the previously defined class $SOLCCP_{\sqcap}(\mathcal{Q})$, but from a user-oriented point of view, $\langle G_{\forall}, K_{\forall} \rangle$ seems the most reasonable solution to $\mathcal{Q}$. Giving up the $\texttt{HomePC}$ concept in $Dem$—and then $(\geq 1 \ \texttt{hasMonitor})$ because of the axiom on $\texttt{HomePC}$—the demander keeps all the specifications on requested components, but they are vacuously true, since $K_{\geq} \sqcap Sup$ implies $\forall \texttt{hasMonitor.}\bot$ *i.e.*, no component is admitted.

In order to make our intuition more precise, we introduce the *number-restriction-minimal* solutions for $\mathcal{Q}$, whose set we denote $SOLCCP_{\mathcal{N}}(\mathcal{Q})$. Intuitively, a solution $\langle G, K \rangle$ for $\mathcal{Q}$ is in $SOLCCP_{\mathcal{N}}(\mathcal{Q})$ when an at-least restriction $(\geq n \ R)$ is in $G$ only if it directly conflicts with an at-most restriction $(\leq m \ R)$ (with $m < n$) in $D$. Solutions in which the at-least restriction is given up because of conflicting universal role quantifications—*e.g.*, $\forall R.A$ and $\forall R.\neg A$—are not in $SOLCCP_{\mathcal{N}}(\mathcal{Q})$. Since this characteristic of number-restriction-minimal solutions should be enforced at any level of nesting, we first introduce the *role path* of a concept in $\mathcal{ALN}$. Here we need to distinguish between a concept $A$ and its (different) occurrences in another concept, *e.g.*, $B = A \sqcap \forall R.A$. In theory, we should mark each occurrence with a number, *e.g.*, $A^1 \sqcap \forall R.A^2$; however, since we need to focus on one occurrence at a time, we just mark it as $\overline{A}$.

**Definition 12** *Given a concept $B$ in $\mathcal{ALN}$, and an occurrence $\overline{A}$ of an atomic (sub)concept $A$ in $B$, a role path for $\overline{A}$ in $B$, $\Pi_{\overline{A}}(B)$ is a string such that:*

- *$\Pi_{\overline{A}}(\overline{A}) = \epsilon$, where $\epsilon$ denotes the empty string*

- *$\Pi_{\overline{A}}(B_1 \sqcap B_2) = \Pi_{\overline{A}}(B_i)$, where $B_i$, $i \in \{1,2\}$, is the concept in which the occurrence of $\overline{A}$ appears*

- *$\Pi_{\overline{A}}(\forall R.B) = R \circ \Pi_{\overline{A}}(B)$, where $\circ$ denotes string concatenation*

The role path $\Pi_{\overline{A}}(B)$ represents the role nesting of a concept $\overline{A}$ occurrence into a concept $B$. Note that $\Pi_{\overline{A}}(B)$ is the same for any commutation of conjunctions in $B$, and for any rearrangement of universal role quantifications—if $A$ was not atomic, this would not be true[7]. Using the previous definition we can now define $SOLCCP_{\mathcal{N}}(\mathcal{Q})$.

**Definition 13** *Let $\mathcal{Q} = \langle \mathcal{ALN}, C, D, \mathcal{T} \rangle$ be a CCP. The set $SOLCCP_{\mathcal{N}}(\mathcal{Q})$ is the subset of solutions $\langle G, K \rangle$ in $SOLCCP_{\sqcap}(\mathcal{Q})$ such that if $(\geq n\ R)$ occurs in $G$ then there exists $(\leq m\ R)$, with $m < n$, occurring in $CNF(D, \mathcal{T})$ and $\Pi_{(\geq n\ R)}(G) = \Pi_{(\leq m\ R)}(CNF(D, \mathcal{T}))$.*

We now illustrate an algorithm *findContract* that returns a solution $\langle G, K \rangle \in SOLCCP_{\mathcal{N}}(\mathcal{Q})$ for $\mathcal{Q} = \langle \mathcal{ALN}, CNF(C, \mathcal{T}), CNF(D, \mathcal{T}), \emptyset \rangle$, that is, it compares two $\mathcal{ALN}$-concepts $C$, and $D$, both already in CNF w.r.t. a TBox $\mathcal{T}$, and computes a number-restriction minimal contraction $\langle G, K \rangle$ of $C$ w.r.t. $D$ *without* considering the TBox.

**Algorithm** *findContract*$(C, D)$;
    **input** $\mathcal{ALN}$ concepts $C$, $D$, both already in CNF
    **output** number-restriction minimal contraction $\langle G, K \rangle$,
        where $\langle G, K \rangle = \langle \top, C \rangle$ means that $C \sqcap D$ is satisfiable
    **variables** concepts $G, K, G', K'$
**begin**
1. **if** $C = \bot$
    **then return** $\langle \bot, \top \rangle$; /* see comment 1 */
2. $G := \top$; $K := \top \sqcap C$; /* see comment 2 */
3. **for each** concept name $A \in K_{names+}$
    **if** there exists a concept $\neg A \in D_{names\neg}$
        **then** $G := G \sqcap A$; delete $A$ from $K$;
4. **for each** concept $(\geq x\ R) \in K_{\sharp}$
    **such that** there is a concept $(\leq y\ R) \in D_{\sharp}$ with $y < x$
        $G := G \sqcap (\geq x\ R)$; delete $(\geq x\ R)$ from $K$;
5. **for each** concept $(\leq x\ R) \in K_{\sharp}$
    **such that** there is a concept $(\geq y\ R) \in D_{\sharp}$ with $y > x$
        $G := G \sqcap (\leq x\ R)$; delete $(\leq x\ R)$ from $K$;
6. **for each** concept $\forall R.F \in K_{all}$
    **if** there exist $\forall R.E \in D_{all}$ **and** (
        **either** $(\geq x\ R) \in K_{\sharp}$ with $x \geq 1$

---

7. For readers that are familiar with the *concept-centered normal form* of concepts (Baader et al., 2003), we note that $\Pi_{\overline{A}}(B)$ is a word for $U_A$ in the concept-centered normal form of $B$.

$\qquad$ **or** $(\geq \ x \ R) \in D_{\sharp}$ with $x \geq 1$ )
$\qquad$ **then let** $\langle G', K' \rangle$ be the result of *findContract*$(F, E)$ **in**
$\qquad\qquad$ $G := G \sqcap \forall R.G'$;
$\qquad\qquad$ replace $\forall R.F$ in $K$ with $\forall R.K'$;
7. **return** $\langle G, K \rangle$;
**end.**

Let us comment on the algorithm:

1. the case in Step 1 cannot occur at the top level, since we assumed $C$ and $D$ be satisfiable in the definition of CCP. However, $\bot$ may occur inside a universal quantification—*e.g.*, $C = \forall R.\bot$—hence, the case of Step 1 may apply in a recursive call of *findContract*, issued from Step 6 of an outer call.

2. in Step 2, the conjunction $\top \sqcap C$ is assigned to $K$ in order to leave $\top$ in $K$ if every other concept is removed by the subsequent steps.

We denote by $\langle G_\emptyset, K_\emptyset \rangle$ solutions for the CCP $\mathcal{Q}_\emptyset = \langle \mathcal{ALN}, CNF(C, \mathcal{T}), CNF(D, \mathcal{T}), \emptyset \rangle$. In this simplified CCP $\mathcal{Q}_\emptyset$, we completely unfold $\mathcal{T}$ in both $C$ and $D$ and then forget it.

**Theorem 4** *The pair $\langle G, K \rangle$ computed by findContract$(C, D)$ is a number-restriction-minimal contraction for $\mathcal{Q}_\emptyset = \langle \mathcal{ALN}, CNF(C, \mathcal{T}), CNF(D, \mathcal{T}), \emptyset \rangle$.*

*Proof.* We first prove that $\langle G, K \rangle$ is a solution for $\mathcal{Q}_\emptyset$, namely, that (i) $G \sqcap K \equiv C$, and that (ii) $K \sqcap D$ is satisfiable. We prove (i) by induction. For the base cases, observe that the claim is true in Step 2 by construction, and that in Steps 3–5 when a conjunct is deleted from $K$, it is also added to $G$. Hence the claim holds when no recursive call is made. For the inductive case, assume the claim holds for each recursive call in Step 6, that is, $G' \sqcap K' \equiv F$ for every concept $\forall R.F \in K_{all}$. Let $G_n, K_n$ be the values of variables $G, K$ before the execution of Step 6, and let $K_n^-$ be the concept $K_n$ without $\forall R.F$. Then, after Step 6 it is:

$$
\begin{aligned}
G \sqcap K = & \quad \text{(by assignment)} \\
G_n \sqcap \forall R.G' \sqcap K_n^- \sqcap \forall R.K' \equiv & \quad \text{(by definition of } \forall) \\
G_n \sqcap K_n^- \sqcap \forall R.(G' \sqcap K') \equiv & \quad \text{(by inductive hypothesis)} \\
G_n \sqcap K_n^- \sqcap \forall R.F \equiv & \quad \text{(by definition of } K_n^-) \\
G_n \sqcap K_n \equiv & \quad \text{(since the base case holds before Step 6)} \\
C &
\end{aligned}
$$

Regarding (ii), the proof is again by induction, where the inductive hypothesis is that $K' \sqcap E$ is satisfiable. Basically, we construct an interpretation $(\Delta, \cdot^{\mathcal{I}})$ with an element $x$ such that $x \in (K \sqcap D)^{\mathcal{I}}$, and show that we can keep constructing $\mathcal{I}$ without contradictions, since contradicting concepts have been deleted from $K$. In the inductive case, we assume the existence of an interpretation $(\Delta', \cdot^{\mathcal{J}})$ for $K' \sqcap E$ such that $y \in \Delta' \cap (K' \sqcap E)^{\mathcal{J}}$, and then build a joint interpretation $(\Delta'', \cdot^{\mathcal{I}''})$ by letting $\Delta'' = \Delta \uplus \Delta'$, $\mathcal{I}'' = \mathcal{I} \cup \mathcal{J} \cup \{\langle x, y \rangle \in R^{\mathcal{I}''}\}$.

We now prove that $\langle G, K \rangle$ is a number-restriction-minimal solution for $\mathcal{Q}_\emptyset$. The proof is by induction on the Quantification Nesting (QN) of $C$, defined in Section 3.1. Observe that an at-least restriction is deleted from $K$ only in Step 4 of *findContract*. For the base case—$QN(C) = 0$, no recursive call—observe that the role path of a retracted concept $(\geq\ n\ R)$ in $G$ is $\epsilon$, same as the role path of the concept $(\leq\ m\ R)$ in $D$ causing Step 4 to be executed. Hence, the claim holds in the base case. For the inductive case, assume that the claim holds for all concepts with QNs smaller than $QN(C)$. Observe that the concept $F$ in Step 6 is such a concept, since its QN is smaller by at least 1. Hence, if an (occurrence of an) at-least restriction $\overline{(\geq\ x\ R)}$, with role path $\Pi_{\overline{(\geq\ x\ R)}}(F)$ is deleted in $F$, there exists a conflicting at-most restriction in $E$ with the same role path. Since both $F$ and $E$ occur inside the scope of a concept $\forall R.F$, $\forall R.E$ respectively, the claim still holds with role path $\Pi_{\overline{(\geq\ x\ R)}}(C) = R \circ \Pi_{\overline{(\geq\ x\ R)}}(F)$. □

## 6.2 Contraction-Based Ranking of Partial Matches

We now define a penalty function $p_{\perp}$ for partial matches based on the following intuition: the partial matches should be ranked based on how many characteristics should be retracted from each $C$ to make them potential matches.

**Algorithm** *penaltyPartial*$(C, D)$;
      **input** $\mathcal{ALN}$ concepts $C$, $D$, both already in CNF
      **output** a penalty for the partial match between $C$ and $D$
          where zero means that $C \sqcap D$ is satisfiable
      **variables** integer $n$
**begin**
1. **if** $C = \perp$
      **then return** $|D|$; /* see Comment 1 */
2.  $n = 0$;
3. **for each** concept name $A \in C_{names+}$
      **if** there exists a concept $\neg A \in D_{names\neg}$
          **then** $n := n + 1$;
4. **for each** concept $(\geq\ x\ R) \in C_{\sharp}$
      **such that** there is a concept $(\leq\ y\ R) \in D_{\sharp}$ with $y < x$
          $n := n + 1$;
5. **for each** concept $(\leq\ x\ R) \in C_{\sharp}$
      **such that** there is a concept $(\geq\ y\ R) \in D_{\sharp}$ with $y > x$
          $n := n + 1$;
6. **for each** concept $\forall R.F \in C_{all}$
      **if** there exist $\forall R.E \in D_{all}$ **and (**
          **either** $((\geq\ x\ R) \in C_{\sharp}$ **and** $(\leq\ y\ R) \notin D_{\sharp}$ with $x \geq y)$ /* see Comment 2 */
          **or** $(\geq\ x\ R) \in D_{\sharp}$ with $x \geq 1$ )
      **then** $n := n + penaltyPartial(F, E)$;
7.  **return** $n$;
**end.**

The above algorithm has a structure very similar to *findContract*: whenever *findContract* removes concepts from $K$, *penaltyPartial* adds penalties to $n$. The two differences are explained in the following comments:

1. Step 1 adds the whole length of $D$ when $C = \bot$. This addition ensures antimonotonicity in the presence of $\bot$, as explained in Example 4 below.

2. Step 6 has in *penaltyPartial* the additional condition "**and** $(\leq y\ R) \notin D_\sharp$ with $x \geq y$". This condition is necessary because *penaltyPartial* does not actually remove concepts, but just counts them. If an at-least restriction in $C_\sharp$ is in contrast with an at-most restriction in $D_\sharp$, then *findContract* removes it from $K$, while *penaltyPartial* just adds 1 to $n$. Yet, when the condition in Step 6 is evaluated, *findContract* finds it false just because the at-least restriction has been removed, while *penaltyPartial* would find it true, were it not for the additional condition.

We now use the outcome of *penaltyPartial* to define a penalty function for partial matches.

**Definition 14** *Given a simple TBox $\mathcal{T}$ in $\mathcal{ALN}$, let the penalty function $p_\bot$ for the partial match of a counteroffer $C$ given an offer $D$, where both $C$ and $D$ are concepts in $\mathcal{ALN}$, be as follows.*

$$p_\bot(C, D, \mathcal{T}) \doteq penaltyPartial(CNF(C, \mathcal{T}), CNF(D, \mathcal{T})) \tag{3}$$

Note that since *penaltyPartial* closely follows *findContract* and *findIrred*, in fact Formula (3) is more similar to Formula (1) in Definition 9 than it might appear. Implicitly, we solve $\mathcal{Q}_\emptyset = \langle \mathcal{ALN}, CNF(C, \mathcal{T}), CNF(D, \mathcal{T}), \emptyset \rangle$, and then use the result in the computation of the penalty function, with a main difference in Step 1, though. We explain such a difference with the help of an example.

**Example 4** Let $Dem_1$ and $Dem_2$ be two demands, where $Dem_2 \sqsubseteq_{\mathcal{T}} Dem_1$, and let $Sup$ be a supply, all modeled using the ontology $\mathcal{T}$ in Figure 1 as in the following:

$$
\begin{aligned}
Dem_1 &= \text{ PC} \sqcap \forall\text{hasMonitor.CRTmonitor} \\
Dem_2 &= \text{ PC} \sqcap \forall\text{hasMonitor.}\bot \\
Sup &= \text{ HomePC} \sqcap \forall\text{hasMonitor.LCDmonitor}
\end{aligned}
$$

Computing *findContract* and *penaltyPartial* for both $CNF(Dem_1, \mathcal{T})$ and $CNF(Dem_2, \mathcal{T})$ w.r.t. $CNF(Sup, \mathcal{T})$ we obtain:

$$
\begin{aligned}
findContract(CNF(Dem_1, \mathcal{T}), CNF(Sup, \mathcal{T})) &= \langle \forall\text{hasMonitor.CRTmonitor}, \\
&\qquad \text{PC} \sqcap \forall\text{hasMonitor.Monitor}\rangle \\
penaltyPartial(CNF(Dem_1, \mathcal{T}), CNF(Sup, \mathcal{T})) &= 1 \\
findContract(CNF(Dem_2, \mathcal{T}), CNF(Sup, \mathcal{T})) &= \langle \forall\text{hasMonitor.}\bot, \text{PC}\rangle \\
penaltyPartial(CNF(Dem_2, \mathcal{T}), CNF(Sup, \mathcal{T})) &= 3
\end{aligned}
$$

In summary, the concept $\bot$ conflicts with every other concept, yet when a concept $\forall R.\bot$ is given up, its length is zero (or any other constant), hence the length of $G$ cannot be directly used as an antimonotonic penalty function. This explains the importance of Step 1 in the above algorithm.

We can show the following formal correspondence between $p_\bot$ and the Concept Contraction defined in the previous Section.

**Theorem 5** *Let $\mathcal{Q} = \langle \mathcal{ALN}, C, D, \mathcal{T} \rangle$ be a CCP, and let $\langle G_\emptyset, K_\emptyset \rangle$ the solution to $\mathcal{Q}_\emptyset$ returned by findContract($CNF(C, \mathcal{T})$, $CNF(D, \mathcal{T})$). If $G_\emptyset$ does not contain any occurrence of the concept $\bot$, then*

$$p_\bot(C, D, \mathcal{T}) = |G_\emptyset|$$

*Proof.* The function $p_\bot$ is based on *penaltyPartial*, and by inspection, whenever *penaltyPartial* increments $n$, *findContract* adds an atomic concept to $G_\emptyset$. The only exception is in Step 1 of *penaltyPartial*, which adds $|D|$ while *findContract* adds $\bot$ to $G_\emptyset$. However, this case is explicitly outside the claim. $\qquad\square$

We now prove that $p_\bot$ is in accordance with properties highlighted in the previous Section.

**Theorem 6** *The penalty function $p_\bot$ is (i) non-symmetric, (ii) syntax independent, and (iii) antimonotonic over subsumption.*

*Proof.* (i) Non-symmetry is proven by example: let $C = (\leq 1\ R) \sqcap \forall R.\neg A$, $D = (\geq 2\ R) \sqcap \forall R.A$. For simplicity, $\mathcal{T} = \emptyset$, and observe that both $C$ and $D$ are already in CNF. We now show that $p_\bot(C, D, \emptyset) \neq p_\bot(D, C, \emptyset)$. In fact, in the former case, observe that $C$ must give up everything: the at-most restriction because it is in contrast with the at-least restriction, and $\neg A$ inside universal quantification because it is in contrast with $\forall R.A$ in $D$. Hence, *penaltyPartial* returns $2 = (1$ from Step 5$) + (1$ from Step 1 of the recursive call$)$. Hence, $p_\bot(C, D, \emptyset) = 2$. In the latter case, instead, once the at-least restriction is given up (and *penaltyPartial* adds 1 to $n$ in Step 4), since role fillers are no more imposed, the universal quantification is now compatible (the condition of the **if** in Step 6 is false). Hence $p_\bot(D, C, \emptyset) = 1$.

(ii) syntax independency is an immediate consequence of the fact that Formula (3) uses normal forms for concepts. Since normal forms are unique up to commutativity of conjunction—that can be fixed by imposing some order to conjunctions, *e.g.*, lexicographic—the claim holds.

(iii) antimonotonicity can be proved by induction on the QN of a generic concept $C'$ subsumed by $C$; we go through all conditions for subsumption, analyzing the changes in the behavior of the algorithm from $C$ to $C'$. Recall that our goal is now to prove that $p_\bot(C', D, \mathcal{T}) \geq p_\bot(C, D, \mathcal{T})$. In order to make a clear distinction between the two computations, we let $n'$ be the (instance of the) variable used in the call to *penaltyPartial*($C', D$), while $n$ is used in the call to *penaltyPartial*($C, D$). To ease notation, we assume that $C, C'$ are already in CNF.

- First of all, it could be the case that $C' = \bot$. In this case, $n' = |D|$ from Step 1 of *penaltyPartial*. On the other hand, observe that *penaltyPartial*($C, D$) $\leq |D|$ because either $C = \bot$ too, or every increase in $n$ corresponds to an atomic concept in $D$—by inspection of Steps 3–5, and this recursively in Step 6. Therefore, the claim holds for this base case.

- $C_{names} \subseteq C'_{names}$. For this case, it is obvious that Step 3 in *penaltyPartial* can only make more increments to $n'$ w.r.t. $n$, since for $C'$ the number of iterations of the **for each** increases.

- for every number restriction in $C_\sharp$, either the same number restriction appears in $C'_\sharp$, or it is strengthened (an at-least increases, an at-most decreases) in $C'_\sharp$. Note that strengthening a number restriction in $C'$ can only turn from false to true the condition for the increment of $n$ in Steps 4–5. For instance, passing from $(\geq \ x \ R) \in C_\sharp$ to $(\geq \ x' \ R) \in C'_\sharp$ with $x' \geq x$, if there is $(\leq \ y \ R) \in D_\sharp$ then $y < x$ implies $y < x'$. A similar argument holds for the at-most. Moreover, number restrictions that appear only in $C'_\sharp$ can only increase the number of iterations of Steps 4–5, hence $n'$ can only increase w.r.t. $n$ and the claim holds.

The above three cases prove the basis of the induction ($C'$ with QN equal to 0). We now prove the case for universal role quantification, assuming that the claim holds for QNs less than $QN(C')$.

- for every $\forall R.F' \in C'_{all}$, either $R$ is not universally quantified in $C_{all}$, or there is $\forall R.F \in C_{all}$ such that $F'$ is subsumed by $F$ (with $F' = F$ as a special case of subsumption). Roles which are not universally quantified in $C_{all}$ but are quantified in $C'_{all}$, can only increase the number of iterations of Step 6, hence $n'$ can only increase due to their presence. For roles that have a more specific restriction $F'$, the inductive hypothesis is assumed to hold, since $QN(F') < QN(C')$. Hence $p_\perp(F', E, \mathcal{T}) \geq p_\perp(F, E, \mathcal{T})$. This is equivalent to $penaltyPartial(F', E) \geq penaltyPartial(F, E)$. Moreover, if the condition in Step 6 is true in the call $penaltyPartial(C, D)$, then it is also true in $penaltyPartial(C', D)$, since $\forall R.F' \in C'_{all}$, and $(\geq \ x' \ R) \in C'_\sharp$, hence if the recursive call $penaltyPartial(F, E)$ is issued, then also $penaltyPartial(F', E)$ is issued, increasing $n'$ at least as much as $n$ is increased, by inductive hypothesis. Hence the claim holds also in the inductive case.

$\square$

## 7. The Matchmaking System

The DLs-based approach to semantic matchmaking illustrated in previous Sections has been implemented in the $\mathcal{ALN}$ reasoning engine MaMaS (MatchMaking Service). It features all classical inference services of a DL reasoner, but also implements algorithms for the non-standard services for matchmaking presented in previous Sections.

MaMaS is a multi-user, multi-ontology Java servlet based system; it is available as an HTTP service at: `http://dee227.poliba.it:8080/MAMAS-tng/DIG`, and exposes a DIG 1.1[8] compliant interface. The basic DIG 1.1 has been extended to cope with non standard services, and we briefly describe here such additions.

New elements:

- Match type detection: `<matchType>E1 E2</matchType>`- computes the match type according to the following classification: Exact (equivalence), Full, Plug-in, Potential, Partial.

---

8. DIG 1.1 is the new standardized DL systems interface developed by the Description Logic Implementation Group (DIG) (Haarslev & Möller, 2003).

- Concept Abduction: `<abduce>E1 E2</abduce>` - implements *findIrred*.

- Concept Contraction: `<contract>E1 E2</contract>`- implements *findContract*.

- Ranking Score: `<rank type="potential">E1 E2</rank>`
  `<rank type="partial">E1 E2</rank>`- computes $p_\sqsubseteq(C, D, \mathcal{T})$ and $p_\perp(C, D, \mathcal{T})$ as presented in previous Sections.

New attributes for `<newKB/>`

- `shared`: the only values to be used are `true` and `false`. In MaMaS, when a new knowledge base is created, each KB uri is associated with the IP address of the client host (owner) instantiating the KB. If the shared attribute is set to `false`, only the owner is authorized to submit tells statements and change the KB as well as to submit asks. In this case, requests from IP addresses different from the owner's one can be only asks. If the shared attribute is set to `true`, then no restriction is set on both tells and asks statements. True is the default value.

- `permanent`: the only values to be used are `true` and `false`. In MaMaS, if a KB is not used for more than 300 seconds, the KB is automatically released. If a user wants to maintain the KB indefinitely, the `permanent` attribute must be set to `true`; `false` is the default value.

It should also be pointed out that MaMaS only supports simple-TBox, that is, concept axioms have a concept name on the left side[9].

We have been using MaMaS as matching engine in various applications, including e-marketplaces, (see *e.g.*, Colucci, Di Noia, Di Sciascio, Donini, Ragone, & Rizzi, 2006; Colucci et al., 2005) and semantic web services discovery (Ragone, Di Noia, Di Sciascio, Donini, Colucci, & Colasuonno, 2007). We do not delve in details of such applications here, and refer the interested reader to the cited references.

## 7.1 Experimental Evaluation

The hypothesis we seek to confirm in this Section is that our approach performs effectively in a wide range of matchmaking scenarios, *i.e.*, it is able to model commonsense human behavior in analyzing and ranking, given a request, available offers. Hence the experimental framework relies on comparison of system behavior versus the judgement of human users. Furthermore, although our system may allow the use of weights to increase the relevance of concepts, in the following results refer to the basic "unweighted" version of the system, to avoid biasing of results due to weights introduction.

The scenarios we tested our approach on were three: apartments rental, date/partner finding, skill management for recruiting agencies. Several ontology design methodologies have been proposed (Jones, Bench-Capon, & Visser, 1998); we adopted the one proposed by N.F. Noy and D.L. McGuinness (2001).

---

9. Notice that since MaMaS supports $\mathcal{ALN}$, only atomic negation can be expressed and then `<disjoint/>` groups must contain only concepts specialized by an `<impliesc>` axiom (sub-concept axiom). Defined concepts `<equalc/>` (same-class) are not admitted in a disjoint group.

For all three scenarios we carried out a thorough domain analysis, starting with a large set of advertisements taken from newspapers or from descriptions of on-line agencies, and designed ontologies describing the domain. In particular:

- Apartments rental ontology is made up of 146 concepts (primitive + defined) and 33 roles.

- Date/partner matching ontology is made up of 131 concepts (primitive + defined) and 29 roles.

- Skill matching ontology is made up of 308 concepts (primitive + defined) and 38 roles.

For each scenario we selected several announcements. The total number used in the experiments with human users is 180 (120 offers, 60 requests) for the apartments rental, 215 (140 offers, 75 requests) for the skill matching. 100 advertisements for the Date matching scenario were also selected, yet for these we did not actually distinguish among requests and offers as announcements were in the form of profiles, although they included preferences for dating partner. All announcements were in natural language and they were manually translated in DL syntax. We then created, for each domain, 50 sets of questionnaires. Questionnaires were in the form of one request (a demand or a supply) and 10 offering advertisements. Three groups of ten randomly selected volunteers, were then asked to order, according to their judgement advertisements, with respect to the given requests. Having obtained average users rankings, we run the same sets of advertisements with our system, which gave us a set of system provided rankings. System rankings that included *partial matching* advertisements were simply ordered below worst *potential matching* advertisement. We adopted, as reference, a standard Vector Space Model (VSM) (Salton & Gill, 1983) system. We used terms in our ontologies "flattening" the ontology descriptions, as dimensions of three separate vector spaces, and determined weights using classical $TF * IDF$ measure. Similarity results were computed using the well-known *Cosine* similarity measure (Salton & Gill, 1983).

To summarize results we adopted the $R_{norm}$ (Bollmann, Jochum, Reiner, Weissmann, & Zuse, 1985) as quality measure of our system effectiveness. $R_{norm}$ is defined as follows. Given $Sup$, a finite set of descriptions with a user-defined preference relation $\geq$ that is complete and transitive, let $\Delta^{usr}$ be the rank ordering of $Sup$ induced by users preference relation, and let $\Delta^{sys}$ be the system-provided ranking. $R_{norm}$ is then defined as:

$$R_{norm}(\Delta^{sys}) = \frac{1}{2} \cdot (1 + \frac{S^+ - S^-}{S_{max}^+})$$

where $S^+$ is the number of descriptions pairs where a better description is ranked by the system ahead of a worse one; $S^-$ is the number of pairs where a worse description is ranked ahead of a better one and $S_{max}^+$ is the maximum possible number of $S^+$. It should be noticed that the calculation of $S^+$, $S^-$, and $S_{max}$ is based on the ranking of descriptions pairs in $\Delta^{sys}$ relative to the ranking of corresponding descriptions pairs in $\Delta^{usr}$. $R_{norm}$ values are in the range [0,1]; a value of 1 corresponds to a system-provided ordering of the available descriptions that is either identical to the one provided by the human users or has a higher degree of resolution, lower values correspond to a proportional disagreement between the two. For the three scenarios considered, results are presented in table 3.

| Domain | MaMaS | VSM |
|---|---|---|
| Apartments rental | 0.87 | 0.48 |
| Date/partner matching | 0.79 | 0.41 |
| Skill matching | 0.91 | 0.46 |

Table 3: $R_{norm}$ values: MaMaS: Semantic matchmaking results, VSM: Vector Space Model results

Although they present a variability, which we believe is partly due to the ability to capture the domain in the ontologies design, they show that our approach provides rankings that are close to human commonsense behavior and are far better than those obtained with unstructured text retrieval tools.

## 8. Conclusion

We have addressed the matchmaking problem between descriptions from a DL perspective. We have analyzed semantic-based matchmaking process and devised general commonsense properties a matchmaker should have. We have also pointed out that classical inference services of DLs, such as satisfiability and subsumption, are needed and useful, but may be not sufficient to cope with challenges posed by matchmaking in an open environment.

Motivated by this we have studied Concept Abduction and Contraction as novel non-monotonic inferences in DLs suitable for modeling semantic-based matchmaking scenarios. We analyzed minimality criteria, and proved simple complexity results. We also presented reasonable algorithms for classifying and ranking matches based on the devised inferences in terms of penalty functions, and proved that they obey to properties individuated.

Although several other measures may be determined to compute a score for "most promising" matches our proposal has logical foundations and we have empyrically shown it is able to well simulate commonsense human reasoning. Obviously, as any other semantic-based approach, also our own has to rely on well-designed ontologies able to model the application domain being considered.

Based on the theoretical work we have implemented a fully functional matchmaking facilitator, oriented to both generic e-marketplace advertisements and to semantic-based web-service discovery, which exploits state of art technologies and protocols, and it is, to the best of our knowledge, the only running system able to cope with Concept Abduction and Concept Contraction problems.

With specific reference to earlier work of the authors on the subject, Di Sciascio et al. (2001) defined matchmaking as satisfiability of concept conjunction. Definitions of potential match and *near-miss i.e.*, partial match, in terms of abduction and belief-revision were outlined, and the need for ranking of matches motivated, in the work of Di Sciascio, Donini, and Mongiello (2002). Di Noia et al. (2003b, 2003c) proposed a semantic-based categorization of matches, logic-based ranking of matches within categories, and properties ranking functions should have, in the framework of E-marketplaces. An extended and revised version of such works is in (Di Noia, Di Sciascio, Donini, & Mongiello, 2004). Di Noia et al. (2003a) intro-duced Concept Abduction in DLs and presented algorithms to solve a Concept Abduction Problem in $\mathcal{ALN}$. Colucci et al. (2003) proposed both Concept Abduction and Concept Contraction as inferences suitable for semantic-matchmaking and explanation services. Calì et al. (2004) proposed a basic approach adopting penalty functions ranking, in the framework of dating systems. Colucci et al. (2004) proposed initial results and algorithms based on truth-prefixed tableau to solve Concept Abduction and Contraction problems in $\mathcal{ALN}$. Colucci et al. (2005) showed that such services can be usefully adopted both for semantic-matchmaking and for finding negotiation spaces in an E-Commerce setting. The use of the proposed inference services for refinement purposes in the semantic-matchmaking process has been outlined in the work of Colucci et al. (2006).

Our current research is oriented to the investigation of algorithms for more expressive DLs and the development of a tableaux-based system for the proposed inference services.

## Acknowledgments

We are grateful to the anonymous reviewers for comments and suggestions that improved the quality of this paper. We thank Andrea Calì and Diego Calvanese for useful discussions, and in particular for suggesting the term "penalty function". Simona Colucci, Azzurra Ragone, Marina Mongiello and all the people at SisInfLab gave us invaluable help and suggestions. This research has been supported by EU FP-6 IST STREP TOWL co. 026896.

## Appendix A. Rules for Normal Form

The normal form of a concept can be obtained by repeatedly applying the rules of the two following Sections, until no rule is applicable at any level of nesting of concepts inside $\forall R.C$.

### A.1 Rules Involving Subconcepts

In the following rules, the $\sqcap$ symbol on the l.h.s. should be considered as an associative and commutative operator; hence, for instance, when writing $(\geq n\ R) \sqcap (\leq m\ R)$ in the second rule, this should be read as the concepts $(\geq n\ R)$ and $(\leq m\ R)$ appear in any order inside a conjunction of two or more concepts.

$$
\begin{aligned}
C \sqcap \bot &\rightarrow \bot \\
(\geq n\ R) \sqcap (\leq m\ R) &\rightarrow \bot \ \text{ if } n > m \\
A \sqcap \neg A &\rightarrow \bot \\
(\geq n\ R) \sqcap (\geq m\ R) &\rightarrow (\geq n\ R) \ \text{ if } n > m \\
(\leq n\ R) \sqcap (\leq m\ R) &\rightarrow (\leq n\ R) \ \text{ if } n < m \\
\forall R.D_1 \sqcap \forall R.D_2 &\rightarrow \forall R.(D_1 \sqcap D_2) \\
\forall R.\bot &\rightarrow \forall R.\bot \sqcap (\leq 0\ R)
\end{aligned}
$$

## A.2 Rules Involving the Concept and the TBox

$$A \rightarrow A \sqcap C \text{ if } A \sqsubseteq C \in \mathcal{T}$$
$$A \rightarrow C \text{ if } A \equiv C \in \mathcal{T}$$
$$A \rightarrow A \sqcap \neg B_1 \sqcap \cdots \sqcap \neg B_k \text{ if } disj(A, B_1, \ldots, B_k) \in \mathcal{T}$$

Usually the concept resulting from the application of the above rules is referred to as an *expansion*, or *unfolding* of a TBox.

## A.3 Properties of the Normal Form

Let $C$ be a concept in Classic, and let $C'$ be any concept obtained from $C$ by repeatedly appying the above rules. Let $|C|, |C'|$ denote the size of $C, C'$ respectively. It can be proved (Borgida & Patel-Schneider, 1994) that:

1. if $|C'|$ is polynomially bounded in $|C|$, then $C'$ can be computed in time $O(|C|^2)$;

2. every concept resulting from the application of the rules is equivalent to $C$, w.r.t. models of the TBox.

As a consequence of the latter property, $C$ is unsatisfiable iff its normal form is $\bot$. Then, as a consequence of the former property, unsatisfiability can be decided in polynomial time (Borgida & Patel-Schneider, 1994). The fact that $|C'|$ is polynomially bounded in $|C|$ has been intuitively related by Nebel (1990) to the form of TBoxes, that should be "bushy but not deep". A more precise definition has been given by Colucci et al. (2004).